\documentclass{article}

\usepackage[main, final]{neurips_2026}

\usepackage[utf8]{inputenc} 
\usepackage[T1]{fontenc}    
\usepackage{hyperref}       
\usepackage{url}            
\usepackage{booktabs}       
\usepackage{amsfonts}       
\usepackage{nicefrac}       
\usepackage{microtype}      
\usepackage[dvipsnames,table]{xcolor} 
\usepackage{graphicx}
\usepackage{multirow}
\definecolor{blue}{RGB}{12,132,198}
\definecolor{red}{RGB}{247,77,77}
\definecolor{green}{RGB}{14,152,111}
\definecolor{purple}{RGB}{121,108,173}
\definecolor{orange}{RGB}{214,88,19}
\definecolor{darkgray}{RGB}{90,90,90}
\usepackage{colortbl}
\usepackage{subcaption}
\title{ShadowFPT: Backdooring Federated Prompt Tuning via Shadow  Triggers}

\author{%
  Kun Zhai\textsuperscript{1} \quad
  Teng Li\textsuperscript{1} \quad
  Yunhao Feng\textsuperscript{2} \quad
  Xingjun Ma\textsuperscript{1}\thanks{Corresponding author: xingjunma@fudan.edu.cn} \\
  \textsuperscript{1}Fudan University, Shanghai, China \\
  \textsuperscript{2}Alibaba Group, Hangzhou, China
}

\begin{document}

\maketitle

\begin{abstract}
Federated Prompt Tuning (FPT) adapts large vision--language models by freezing the pretrained backbone and optimizing only lightweight prompt parameters across clients. Although this design improves communication and parameter efficiency, it creates an overlooked security risk. Since the backbone is shared and frozen, a malicious client can induce backdoor behavior through the representation space while keeping its uploaded prompt updates close to benign ones. We propose \textbf{ShadowFPT}, a targeted backdoor attack that exploits this frozen-backbone attack surface. ShadowFPT first pretrains a learnable \emph{Shadow Trigger} against the frozen CLIP visual encoder, using either auxiliary public data or the malicious client's local data, so that triggered inputs are steered toward the target class in representation space. During federated prompt tuning, the malicious client adapts the trigger under the current global prompt and then optimizes its local prompt on both clean and triggered samples. Only prompt parameters are uploaded to the server, while the trigger and frozen encoders remain local. By shifting most of the attack burden from prompt updates to trigger-induced representation steering, ShadowFPT achieves targeted misclassification while preserving prompt-space stealthiness. Across multiple datasets, aggregation rules, and non-IID federated partitions, ShadowFPT increases the attack success rate from 19.81\% to 90.36\% in our main setting, while maintaining clean accuracy. It remains effective across textual, visual, and joint vision--language prompt tuning. These results identify frozen backbones as stealthy and underexplored backdoor surfaces in federated prompt tuning, suggesting that defenses based only on prompt-update anomaly detection are insufficient.
\end{abstract}

\section{Introduction}
\label{sec:intro}
Large vision--language models (VLMs), such as CLIP~\cite{radford2021learning}, learn transferable image--text representations and have become a standard foundation for multimodal recognition and adaptation. Adapting these models in distributed environments, however, remains difficult because full fine-tuning of large pretrained backbones is computationally expensive and communication-intensive. Prompt tuning addresses these limitations by keeping the pretrained backbone fixed and optimizing only a small set of learnable prompt parameters~\cite{khattak2023maple}. This parameter-efficient formulation is particularly well suited to federated learning. In \emph{Federated Prompt Tuning} (FPT), clients update lightweight prompts using private local data and communicate only prompt parameters to the server, while the VLM backbone remains frozen.

Backdoor attacks remain a persistent security threat in federated learning. Classical attacks, such as BadNets~\cite{gu2017badnets} and Blended~\cite{chen2017targeted}, poison training data with attacker-designed triggers, causing the trained model to retain high clean accuracy while predicting an attacker-specified target label on triggered inputs. In federated learning, malicious clients can further inject backdoors through their local model updates, making attack persistence under server aggregation a central challenge. Prior work has studied this problem in standard and personalized federated learning, including A3FL~\cite{zhang2023a3fl} and Bad-PFL~\cite{fan2025bad}. Backdoor risks have also been examined in vision--language models and prompt learning. For example, BadCLIP attacks multimodal contrastive learning by guiding both visual and textual embeddings~\cite{liang2024badclip}. However, backdoor attacks against federated prompt tuning remain insufficiently understood. In particular, existing studies largely overlook the security risk that arises when the server aggregates only lightweight prompt parameters while all clients rely on a shared frozen VLM backbone.

Most federated backdoor defenses operate on the updates communicated by clients. Representative methods identify updates that are close to the client majority~\cite{blanchard2017machine}, remove coordinate-wise outliers~\cite{yin2018byzantine}, downweight clients with highly similar update directions~\cite{fung2018mitigating}, calibrate client updates using trusted server data~\cite{cao2020fltrust}, or combine clustering, clipping, noise injection, and certification to improve robustness~\cite{nguyen2022flame}. Recent work extends this update-centered view to federated prompt learning. For example, SABRE-FL detects malicious clients through anomaly analysis in embedding space~\cite{khan2025sabre}. Despite their methodological differences, these defenses share a common assumption that backdoor behavior will manifest as detectable anomalies in uploaded updates or in representations derived from them.

This assumption becomes fragile in FPT. The server observes only low-dimensional prompt updates, while the dominant representation geometry is determined by the shared frozen VLM backbone. A malicious client can therefore shift much of the backdoor effect to the interaction between the trigger and the frozen representation space, while keeping its uploaded prompt updates close to benign ones. This creates a security gap between where the backdoor behavior is realized and what the server can observe. The same frozen backbone that enables efficient adaptation also provides a stable and underexplored attack surface for stealthy backdoor injection.

Motivated by this observation, we propose \textbf{ShadowFPT}, a targeted backdoor attack for federated prompt tuning that exploits the shared frozen CLIP visual encoder as a representation-level attack surface rather than relying on conspicuous prompt updates. ShadowFPT consists of two stages. First, it pretrains a learnable \emph{Shadow Trigger} using either auxiliary public data or the malicious client's local data. The trigger is optimized to steer triggered images toward the target class in the frozen CLIP representation space by aligning final image embeddings with the target text anchor and intermediate visual features with target-class layer-wise prototypes. Second, during federated training, the malicious client adapts the Shadow Trigger under the current global prompt and then optimizes its local prompt on both clean and triggered samples to implant the targeted backdoor. Only prompt parameters are uploaded to the server, while the trigger and frozen encoders remain local. By shifting much of the attack effect from uploaded prompt updates to trigger-induced representation steering, ShadowFPT achieves targeted misclassification while keeping malicious prompt updates close to benign updates.

We evaluate ShadowFPT across multiple datasets, aggregation rules, and non-IID partitions. In the main setting, ShadowFPT improves attack success by 70.55 percentage points while largely preserving clean accuracy. It remains effective across textual, visual, and joint vision--language prompt tuning, and prompt-space analyses show that malicious updates stay close to benign ones. These results reveal frozen backbones as stealthy backdoor surfaces and challenge defenses based solely on prompt-update anomaly detection.

The main contributions of this paper are as follows:
\begin{itemize}
    \item We identify a frozen-backbone backdoor surface in FPT, showing that targeted backdoor behavior can be realized through trigger-induced representation steering while malicious prompt updates remain close to benign updates.

    \item We propose \textbf{ShadowFPT}, a targeted backdoor attack that shifts the attack burden from conspicuous prompt-space poisoning to representation-level manipulation of the shared frozen backbone.

    \item We introduce the \emph{Shadow Trigger}, a target-directed trigger learned through target-text alignment and layer-wise visual prototype alignment, enabling triggered images to move toward the target class in the frozen representation space.

    \item We evaluate ShadowFPT across textual, visual, and joint vision--language prompt tuning under non-IID federated settings, demonstrating high attack success, minimal clean-accuracy degradation, and strong prompt-space stealthiness.
\end{itemize}

\section{Related Work}
\label{sec:related}

\noindent\textbf{Federated Prompt Tuning.}
Vision--language models (VLMs), such as CLIP, learn transferable image--text representations and provide strong foundations for multimodal adaptation~\cite{radford2021learning}. Prompt tuning adapts these models efficiently by freezing the pretrained backbone and optimizing only a small set of prompt parameters. Representative CLIP-based approaches include textual prompt tuning~\cite{zhou2022learning,zhou2022conditional}, visual prompt tuning~\cite{jia2022visual}, joint vision--language prompt learning~\cite{khattak2023maple}, and semantic regularization~\cite{khattak2023self}. This parameter-efficient paradigm naturally extends to federated learning, where clients communicate prompt parameters rather than full model updates. Recent studies on federated prompt tuning investigate cooperative prompt learning~\cite{guo2023promptfl}, personalization~\cite{guo2023pfedprompt}, text-driven prompt generation~\cite{qiu2024federated}, personalization--generalization trade-offs~\cite{cui2024harmonizing}, mixture-of-experts prompts~\cite{luo2025mixture}, and multimodal visual prompt tuning~\cite{singha2025fedmvp}. These works primarily focus on communication efficiency, personalization, and generalization under heterogeneous client data. In contrast, the security risks of federated prompt tuning, especially its vulnerability to backdoor attacks, remain insufficiently understood.

\noindent\textbf{Backdoor Attacks.}
Backdoor attacks aim to make a model behave normally on clean inputs while predicting an attacker-specified target label on inputs containing a trigger. Classical attacks achieve this behavior by poisoning training data with visible or blended triggers~\cite{gu2017badnets,chen2017targeted}. In federated learning, the backdoor must also persist through local training and server aggregation, which has motivated attacks based on stealthy poisoned updates, persistent backdoor injection, adaptive optimization, collusive trigger design, and personalized federated learning~\cite{lyu2023poisoning,zhang2022neurotoxin,zhang2023a3fl,lyu2024coba,fan2025bad}.Backdoor vulnerabilities have also been studied in vision--language models. For example, BadCLIP manipulates multimodal contrastive representations through visual and textual embedding guidance~\cite{liang2024badclip}. Existing attacks, however, are largely built around data poisoning, model-update manipulation, or direct prompt-parameter poisoning. FPT presents a different regime in which both the trainable and communicated parameters are small, and local optimization is lightweight. These constraints leave limited capacity for conventional backdoor injection through uploaded updates, motivating our investigation of a more concealed representation-level attack surface.

\noindent\textbf{Backdoor Defenses.}
Federated defenses against malicious clients have largely followed the paradigm of Byzantine-robust aggregation. Classical methods, such as Krum~\cite{blanchard2017machine}, TrimmedMean~\cite{yin2018byzantine}, and Bulyan~\cite{guerraoui2018hidden}, mitigate malicious behavior by suppressing abnormal client updates through robust statistics and outlier filtering. Subsequent approaches improve robustness under heterogeneous or unreliable clients using geometric aggregation, trust bootstrapping, or sparsification~\cite{pillutla2022robust,karimireddy2022byzantine,cao2020fltrust,xu2025achieving}. Defenses designed specifically for federated backdoors further target poisoned global models. FLAME combines client clustering with model clipping~\cite{nguyen2022flame}, while CRFL provides certified robustness against federated backdoors~\cite{xie2021crfl}. Recent studies also investigate secure aggregation and detectability, privacy-constrained robustness under heterogeneity, system- and optimization-level mitigation, and dynamic behavior-based identification~\cite{huang2025fedid,zhang2025runtime}.

Despite their methodological differences, these defenses largely share the assumption that malicious behavior will appear as detectable abnormalities in uploaded updates or in representations derived from them. This assumption becomes fragile in FPT. Under non-IID data, benign prompt updates may vary substantially across clients, while the backdoor effect can be shifted into the interaction between the trigger and the shared frozen representation space rather than being exposed through conspicuous prompt updates. This gap motivates our investigation of frozen-backbone representation spaces as stealthy backdoor surfaces in federated prompt tuning.

\section{Method}
\label{sec:method}

\subsection{Problem Setup and Threat Model}
\label{subsec:setup}

\noindent\textbf{Federated Prompt Tuning.}
We consider federated prompt tuning of a CLIP model composed of a visual encoder $f_I$ and a text encoder $f_T$, both of which remain frozen throughout training. The system contains $N$ clients, where client $i$ holds a private local dataset $\mathcal{D}_i=\{(x_j,y_j)\}_{j=1}^{n_i}$ drawn from a potentially non-IID distribution. Let $\mathcal{C}$ denote the downstream label set. For clarity, we first present the formulation for textual prompt tuning.

Following standard CLIP prompt tuning, each class $c\in\mathcal{C}$ is represented by a prompted text input, such as ``a photo of a [CLASS]'', where the hand-crafted context tokens are replaced by learnable continuous tokens. We denote the resulting prompted text feature by $f_T(c;P)$. Given an image $x$ and prompt parameters $P$, the prompted CLIP classifier computes the class logit as
\begin{equation}
    \ell_c(x;P)
    =
    \left\langle
    f_I(x), f_T(c;P)
    \right\rangle,
    \qquad c\in\mathcal{C},
\end{equation}
where $f_I(x)$ and $f_T(c;P)$ are the image and prompted text features, respectively, and $\langle \cdot,\cdot \rangle$ denotes cosine similarity. We write $F(x;P)$ for the resulting prompted classifier.

At communication round $t$, the server broadcasts the current global prompt $P^t$ to a subset of clients $\mathcal{S}_t$. Each selected client initializes its local prompt with $P^t$ and optimizes the standard classification objective:
\begin{equation}
    \mathcal{L}(P_i)
    =
    \mathbb{E}_{(x,y)\sim\mathcal{D}_i}
    \left[
    \mathrm{CE}\big(F(x;P_i),y\big)
    \right],
\end{equation}
where $\mathrm{CE}(\cdot,\cdot)$ denotes the cross-entropy loss. After local optimization, clients upload their prompt parameters to the server, which aggregates the received prompts using an aggregation rule $\mathrm{Agg}(\cdot)$:
\begin{equation}
    P^{t+1}
    =
    \mathrm{Agg}\left(
    \{P_i^{t+1}: i\in\mathcal{S}_t\}
    \right).
\end{equation}

\noindent\textbf{Threat Model.}
We consider a targeted backdoor attack against FPT with a frozen vision--language backbone. The adversary aims to preserve the clean accuracy of the global model while causing triggered inputs, $x^\delta=\mathcal{A}(x,\delta)$, to be classified as an attacker-specified target class $y_t$. The malicious client has access to local target-class samples, denoted by $\mathcal{R}_{y_t}$, which are used to guide target-directed trigger optimization. During local training, the malicious client can optimize both its local prompt parameters and the trigger $\delta$, but uploads only prompt parameters to the server. The adversary cannot modify the frozen image encoder, the frozen text encoder, the global training protocol, or the server-side aggregation rule. We assume a white-box threat setting in which the malicious client knows the frozen backbone architecture.

\begin{figure}[!htbp]
  \centering
  \includegraphics[width=0.9\linewidth]{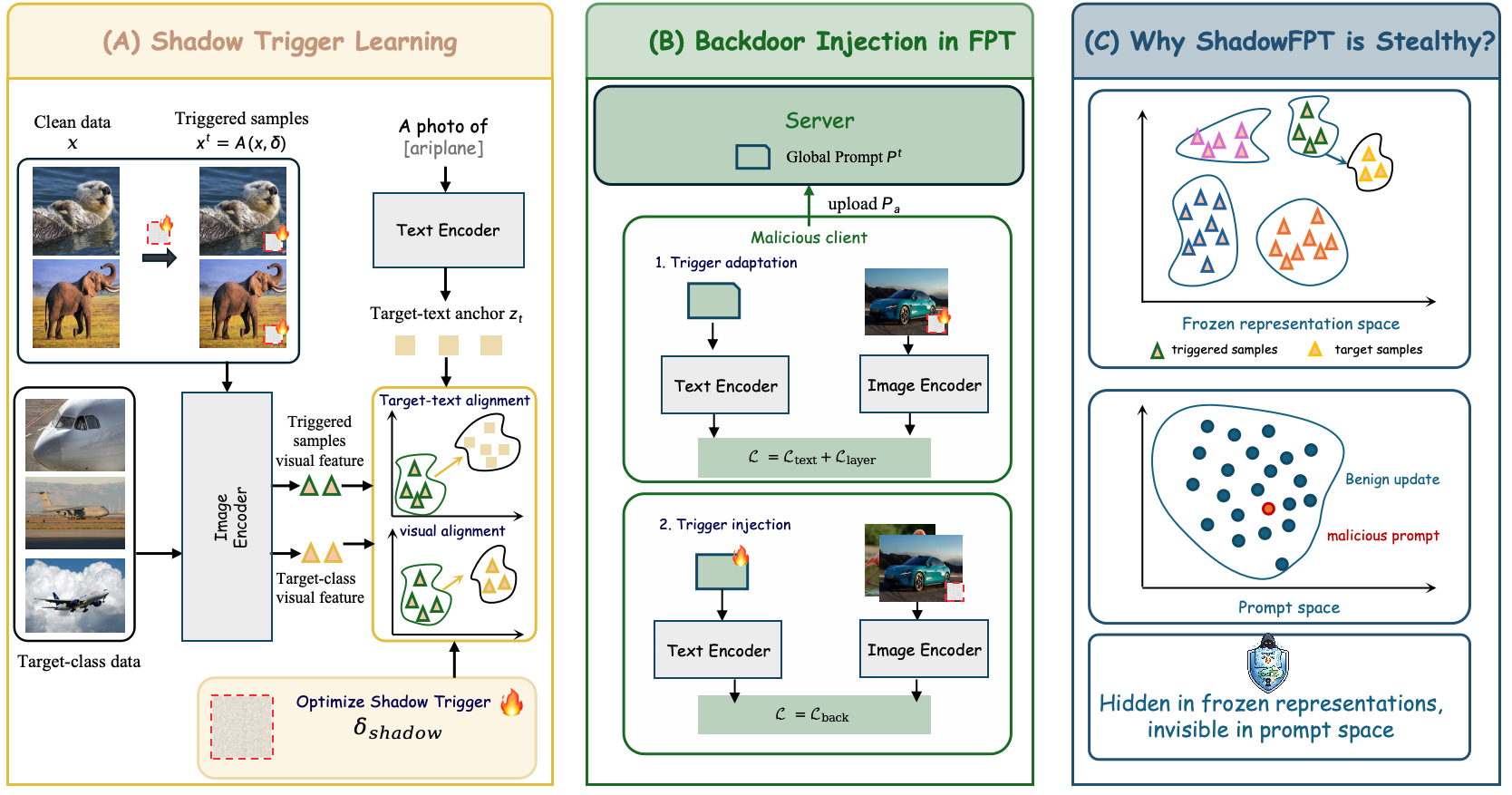}
\caption{
Overview of ShadowFPT. 
(A) The adversary pretrains a Shadow Trigger $\delta_{\mathrm{shadow}}$ against the frozen CLIP backbone through target-text and layer-wise visual alignment. 
(B) During FPT, the malicious client adapts the trigger under the current global prompt $P^t$ and optimizes its local prompt on clean and triggered samples, while uploading only prompt parameters. 
(C) By shifting the attack effect to trigger-induced representation steering, ShadowFPT achieves targeted misclassification with benign-like prompt updates.
}
  \label{fig:overview}
\end{figure}

\subsection{Overview of ShadowFPT}
\label{subsec:overview}

ShadowFPT backdoors FPT through a two-stage procedure. It first learns a \emph{Shadow Trigger} against the frozen CLIP representation space, and then uses this trigger to implant targeted backdoor behavior during malicious local prompt training. Figure~\ref{fig:overview} provides an overview of the ShadowFPT.

Before federated training, the adversary optimizes a learnable trigger $\delta$ on a pretraining dataset $\mathcal{D}_{\mathrm{s}}$, which can be either auxiliary public data or the malicious client's local data. The optimized trigger, denoted by $\delta_{\mathrm{shadow}}$, is trained to steer triggered images toward the target class in the frozen CLIP representation space, as described in Section~\ref{subsec:trigger_pretraining}.

During federated training, the malicious client initializes the trigger with $\delta_{\mathrm{shadow}}$ and adapts it under the current global prompt, allowing the trigger to remain aligned with the evolving federated prompt. The adapted trigger is then applied to a subset of the malicious client's local data, and the local prompt is optimized on both clean and triggered samples so that triggered inputs are classified as the target class. After local optimization, the malicious client uploads only the prompt parameters to the server, while the trigger remains local. This injection procedure is detailed in Section~\ref{subsec:federated_injection}.

\subsection{Shadow Trigger Pretraining}
\label{subsec:trigger_pretraining}

ShadowFPT learns a \emph{Shadow Trigger} by optimizing a learnable trigger $\delta$ against the frozen CLIP visual encoder. Let $\mathcal{D}_{\mathrm{s}}$ denote the data used for trigger pretraining, which can be either auxiliary public data or the malicious client's local data. For an image $x\sim\mathcal{D}_{\mathrm{s}}$, the triggered image is defined as $x^\delta=\mathcal{A}(x,\delta)$. During this stage, only the trigger $\delta$ is updated, while both the visual encoder $f_I$ and the text encoder $f_T$ remain fixed. The goal is to obtain a trigger that steers triggered images toward the target class in the frozen representation space.

\noindent\textbf{Target-text alignment.}
During Shadow Trigger pretraining, we use fixed hand-crafted CLIP templates instead of the learnable federated prompt. For each class $c\in\mathcal{C}$, we construct a text anchor using the frozen text encoder:
\begin{equation}
    z_c
    =
    \frac{1}{|\mathcal{T}|}
    \sum_{T\in\mathcal{T}}
    f_T(T(c)),
\end{equation}
where $\mathcal{T}$ is a fixed set of templates, such as ``a photo of a [CLASS]''. Let $z_t$ denote the text anchor of the target class $y_t$. We optimize the trigger so that the image feature of each triggered input is aligned with $z_t$:
\begin{equation}
    \mathcal{L}_{\mathrm{text}}
    =
    \mathbb{E}_{x\sim\mathcal{D}_{\mathrm{s}}}
    \left[
    1-
    \left\langle
    f_I(x^\delta), z_t
    \right\rangle
    \right],
\end{equation}
where $\langle\cdot,\cdot\rangle$ denotes cosine similarity. This loss encourages the Shadow Trigger to steer triggered images toward the target-class semantic region in the frozen CLIP space.

\noindent\textbf{Layer-wise visual alignment.}
Target-text alignment constrains the final image--text similarity, but it does not explicitly control how the trigger changes intermediate visual representations. We therefore align triggered features with target-class prototypes within the frozen visual encoder. For each selected visual layer $l\in\mathcal{K}$, we compute a target-class prototype from the target-class image set $\mathcal{R}_{y_t}$:
\begin{equation}
    \mu_l^{y_t}
    =
    \frac{1}{|\mathcal{R}_{y_t}|}
    \sum_{x\in\mathcal{R}_{y_t}}
    h_l(x),
\end{equation}
where $h_l(x)$ denotes the layer-$l$ visual feature. The layer-wise alignment loss is defined as
\begin{equation}
    \mathcal{L}_{\mathrm{layer}}
    =
    \mathbb{E}_{x\sim\mathcal{D}_{\mathrm{s}}}
    \left[
    \frac{1}{|\mathcal{K}|}
    \sum_{l\in\mathcal{K}}
    \left(
    1-
    \left\langle
    h_l(x^\delta),
    \mu_l^{y_t}
    \right\rangle
    \right)
    \right].
\end{equation}
This term encourages triggered images to follow the target-class representation trajectory through the frozen visual backbone.

\noindent\textbf{Objective.}
The Shadow Trigger is pretrained by minimizing
\begin{equation}
    \mathcal{L}_{\mathrm{shadow}}
    =
    \mathcal{L}_{\mathrm{text}}
    +
    \lambda_l \mathcal{L}_{\mathrm{layer}}.
\end{equation}
The optimized trigger is denoted by $\delta_{\mathrm{shadow}}$ and is used to initialize the malicious client's trigger during federated prompt tuning.

\subsection{Backdoor Injection in FPT}
\label{subsec:federated_injection}

During federated training, ShadowFPT uses the pretrained Shadow Trigger to implant targeted backdoor behavior into the federated prompt. At communication round $t$, the malicious client receives the current global prompt $P^{t}$ from the server and initializes its local prompt as $P_a \leftarrow P^{t}$. It then performs two sequential local steps. First, it adapts the trigger under the current global prompt. Second, it optimizes the local prompt on poisoned local data.

\noindent\textbf{Trigger adaptation.}
In the first attack round, the malicious client initializes the trigger with $\delta_{\mathrm{shadow}}$ and then updates it locally across subsequent rounds. At round $t$, the prompt is fixed to the current global prompt $P^{t}$, and only the trigger is optimized. We construct the target text anchor under the current global prompt as:
$z_{\mathrm{target}}^{(t)} = f_T(y_t;P^{t})$.
The trigger is adapted using the same representation-level objective as in Shadow Trigger pretraining:
\begin{equation}
    \mathcal{L}_{\mathrm{adapt}}^{(t)}
    =
    \mathcal{L}_{\mathrm{text}}^{(t)}
    +
    \lambda_l \mathcal{L}_{\mathrm{layer}}^{(t)}.
\end{equation}
Here, $\mathcal{L}_{\mathrm{text}}^{(t)}$ aligns triggered image embeddings with the current target text anchor $z_{\mathrm{target}}^{(t)}$, while $\mathcal{L}_{\mathrm{layer}}^{(t)}$ keeps intermediate visual features close to the target-class visual prototypes defined in Section~\ref{subsec:trigger_pretraining}. This step produces a prompt-adapted trigger $\delta^{(t)}$.

\noindent\textbf{Backdoor prompt optimization.}
With the adapted trigger $\delta^{(t)}$ fixed, the malicious client optimizes its local prompt $P_a$ using both clean and triggered local samples. Let $\mathcal{D}_a$ denote the malicious client's local dataset. The prompt-level backdoor objective is
\begin{equation}
    \mathcal{L}_{\mathrm{back}}(P_a)
    =
    \mathbb{E}_{(x,y)\sim\mathcal{D}_{a}}
    \left[
    \mathrm{CE}\big(F(x;P_a),y\big)
    \right]
    +
    \mathbb{E}_{(x,y)\sim\mathcal{D}_{a},\,y\neq y_t}
    \left[
    \mathrm{CE}\big(F(x^{\delta^{(t)}};P_a),y_t\big)
    \right].
\end{equation}
The first term preserves clean-task performance, whereas the second term implants the targeted backdoor into the local prompt.

After local optimization, the malicious client uploads only the prompt parameters $P_a$ to the server. The adapted trigger $\delta^{(t)}$ and the frozen CLIP encoders are never communicated. The server then aggregates the uploaded prompts using the same aggregation rule as in standard FPT.

\section{Experiments}
\label{sec:experiments}

\subsection{Experimental Setup}

\noindent\textbf{Federated Setting.}
All methods use a frozen CLIP model with ViT-B/16 as the backbone. Unless otherwise specified, we use textual prompt tuning as the default FPT setting, denoted by FPT-L~\cite{li2024one}. To evaluate whether ShadowFPT transfers across different prompt parameterizations, we also consider visual prompt tuning (FPT-V)~\cite{jia2022visual} and joint vision--language prompt tuning (FPT-VL)~\cite{khattak2023maple}. Following FedTPG~\cite{qiu2024federated}, we simulate a non-IID federation with 20 clients. Each client holds samples from five classes, and the class sets are disjoint across clients. All clients participate in every communication round.

Unless otherwise stated, a single client is malicious, and the target class is set to the first class of each dataset. All methods are trained for 20 communication rounds with one local epoch per round, batch size 32, and learning rate $0.0035$ for local prompt optimization. We report clean accuracy (Acc) on benign test samples and attack success rate (ASR) on triggered non-target test samples.

\noindent\textbf{ShadowFPT Variants.}
We evaluate two variants of ShadowFPT that differ in the data used for Shadow Trigger pretraining. \textbf{ShadowFPT-I} pretrains the Shadow Trigger using only the malicious client's local data, whereas \textbf{ShadowFPT-II} uses auxiliary public data, instantiated as 100 randomly sampled ImageNet classes. We do not assume semantic overlap between the auxiliary ImageNet classes and the downstream target classes. Both variants use a $32\times 32$ trigger patch placed at the bottom-right corner of the image. The Shadow Trigger is optimized with learning rate $0.01$ during pretraining and $0.001$ during federated trigger adaptation.

\noindent\textbf{Datasets.}
We evaluate ShadowFPT and all baselines on Caltech101~\cite{fei2004learning}, Oxford Flowers~\cite{nilsback2008automated}, and UCF101~\cite{soomro2012ucf101}. These datasets cover generic object recognition, fine-grained flower classification, and human action recognition, respectively.

\noindent\textbf{Attack Baselines.}
We compare ShadowFPT with three groups of attack baselines. The first group includes generic backdoor attacks, namely \textbf{BadNets}~\cite{gu2017badnets} and \textbf{Blended}~\cite{chen2017targeted}. The second group includes federated backdoor attacks, namely \textbf{Neurotoxin}~\cite{zhang2022neurotoxin}, \textbf{A3FL}~\cite{zhang2023a3fl}, \textbf{FCBA}~\cite{liu2024beyond}, and \textbf{CerP}~\cite{lyu2023poisoning}. The third group includes personalized federated backdoor attacks, namely \textbf{Bad-PFL}~\cite{fan2025bad} and \textbf{PFedBA}~\cite{lyu2024lurking}. 

\noindent\textbf{Defense Baselines.}
We evaluate defenses covering several aggregation and filtering mechanisms. The first category is standard aggregation, represented by \textbf{FedAvg}~\cite{mcmahan2017communication}. The second category is Byzantine-robust aggregation, including \textbf{Krum}~\cite{blanchard2017machine}, \textbf{TrimmedMean}~\cite{yin2018byzantine}, and \textbf{Bucketing}~\cite{karimireddy2022byzantine}. The third category includes adaptive or similarity-based filtering methods, namely \textbf{LaSA}~\cite{xu2025achieving} and \textbf{FoolsGold}~\cite{fung2018mitigating}. We also evaluate trust-based aggregation with \textbf{FLTrust}~\cite{cao2020fltrust}, as well as backdoor-specific defenses, including \textbf{FLAME}~\cite{nguyen2022flame} and \textbf{CRFL}~\cite{xie2021crfl}.

\subsection{Main Results}

\begin{table}[!htbp]
\centering
\vspace{-0.3cm}
\caption{
Overall attack performance across datasets and prompt-tuning settings. Results are averaged over all aggregation and defense rules. Best Base. denotes the strongest non-ShadowFPT baseline under each dataset-setting pair, selected by the highest average ASR. $\Delta$Acc and $\Delta$ASR report absolute differences from Best Base., measured in percentage points.
}
\label{tab:overall_main}
\scriptsize
\setlength{\tabcolsep}{3.5pt}
\resizebox{\linewidth}{!}{
\begin{tabular}{ll l cc cccc cccc}
\toprule
\multirow{2}{*}{Dataset}
& \multirow{2}{*}{Setting}
& \multirow{2}{*}{Best Base}
& \multicolumn{2}{c}{Baseline}
& \multicolumn{4}{c}{ShadowFPT-I}
& \multicolumn{4}{c}{ShadowFPT-II} \\
\cmidrule(lr){4-5}
\cmidrule(lr){6-9}
\cmidrule(lr){10-13}
& & & Acc & ASR
& Acc & ASR & $\Delta$Acc & $\Delta$ASR
& Acc & ASR & $\Delta$Acc & $\Delta$ASR \\
\midrule
\multirow{3}{*}{Caltech101}
& FPT-L  & A3FL   & 88.49 & 19.81 & 88.32 & 84.28 & -0.17 & +64.47 & 87.67 & \textbf{90.36} & -0.82 & \textbf{+70.55} \\
& FPT-V  & PFedBA & 92.73 & 37.84 & 92.62 & 92.93 & -0.11 & +55.09 & 91.68 & \textbf{94.25} & -1.05 & \textbf{+56.41} \\
& FPT-VL & PFedBA & 92.82 & 51.39 & 92.39 & 93.19 & -0.43 & +41.80 & 92.28 & \textbf{98.03} & -0.54 & \textbf{+46.64} \\
\midrule
\multirow{3}{*}{Oxford Flowers}
& FPT-L  & A3FL   & 66.63 & 38.07 & 60.93 & 96.74 & -5.70 & +58.67 & 60.85 & \textbf{98.93} & -5.78 & \textbf{+60.86} \\
& FPT-V  & PFedBA & 65.73 & 30.22 & 66.60 & 98.37 & +0.87 & +68.15 & 66.61 & \textbf{98.94} & +0.88 & \textbf{+68.72} \\
& FPT-VL & PFedBA & 69.89 & 54.60 & 68.52 & 98.07 & -1.37 & +43.47 & 68.33 & \textbf{99.18} & -1.56 & \textbf{+44.58} \\
\midrule
\multirow{3}{*}{UCF101}
& FPT-L  & PFedBA  & 68.02 & 25.06 & 66.05 & 99.81 & -1.97 & +74.75 & 65.87 & \textbf{99.98} & -2.15 & \textbf{+74.92} \\
& FPT-V  & PFedBA  & 65.76 & 31.55 & 65.28 & 98.04 & -0.48 & +66.49 & 65.21 & \textbf{99.69} & -0.55 & \textbf{+68.14} \\
& FPT-VL & Blended & 70.14 & 22.49 & 67.97 & 99.28 & -2.17 & +76.79 & 67.89 & \textbf{99.63} & -2.25 & \textbf{+77.14} \\
\bottomrule
\end{tabular}}

\end{table}

\noindent\textbf{Overall effectiveness.}
We first evaluate whether ShadowFPT can implant targeted backdoors while preserving clean-task performance. Table~\ref{tab:overall_main} reports results averaged over all aggregation and defense rules, with detailed results for all datasets, baselines, and defenses provided in Appendix~\ref{app}. Because Best Base denotes the strongest non-ShadowFPT attack for each dataset-setting pair, this comparison is conservative. ShadowFPT consistently outperforms Best Base across all datasets and prompt-tuning settings. Averaged over the nine settings, Best Base achieves 34.56\% ASR, whereas ShadowFPT-I and ShadowFPT-II reach 95.63\% and 97.67\%, respectively. ShadowFPT-II is the strongest method in every setting, improving ASR by 44.58 to 77.14 percentage points.

These gains are achieved with limited degradation in clean accuracy. On Caltech101, ShadowFPT-II reduces clean accuracy by at most 1.05 points. On UCF101, the drop ranges from 0.55 to 2.25 points despite near-100\% ASR. Oxford Flowers under FPT-L shows a larger drop of 5.78 points, suggesting a stronger clean-attack trade-off in fine-grained recognition. This trade-off is much weaker under FPT-V and FPT-VL, where ShadowFPT-II maintains comparable clean accuracy while achieving 98.94\% and 99.18\% ASR. Overall, these results show that ShadowFPT substantially improves targeted attack success without primarily relying on degrading benign performance.

\begin{table}[!htbp]
\centering
\vspace{-0.3cm}
\caption{
Robustness of ShadowFPT against aggregation and defense rules. Results are averaged over all datasets and prompt-tuning settings. 
}
\label{tab:defense_robustness}
\scriptsize
\setlength{\tabcolsep}{3.5pt}
\resizebox{\linewidth}{!}{
\begin{tabular}{l l cc cccc cccc}
\toprule
\multirow{2}{*}{Defense}
& \multirow{2}{*}{Best Base}
& \multicolumn{2}{c}{Baseline}
& \multicolumn{4}{c}{ShadowFPT-I}
& \multicolumn{4}{c}{ShadowFPT-II} \\
\cmidrule(lr){3-4}
\cmidrule(lr){5-8}
\cmidrule(lr){9-12}
& & Acc & ASR
& Acc & ASR & $\Delta$Acc & $\Delta$ASR
& Acc & ASR & $\Delta$Acc & $\Delta$ASR \\
\midrule
FedAvg
& A3FL
& 77.91 & 60.93
& 75.56 & 96.59 & -2.35 & +35.66
& 75.67 & \textbf{98.30} & -2.24 & \textbf{+37.37} \\

Krum
& PFedBA
& 73.30 & 22.66
& 71.70 & 95.60 & -1.60 & +72.94
& 72.20 & \textbf{96.63} & -1.09 & \textbf{+73.97} \\

TrimmedMean
& PFedBA
& 78.08 & 32.64
& 75.50 & 96.11 & -2.58 & +63.47
& 75.55 & \textbf{98.03} & -2.54 & \textbf{+65.38} \\

LaSA
& PFedBA
& 77.16 & 21.34
& 75.72 & 96.16 & -1.44 & +74.82
& 75.71 & \textbf{97.90} & -1.45 & \textbf{+76.55} \\

FoolsGold
& PFedBA
& 73.99 & 39.96
& 75.99 & 95.83 & +2.01 & +55.87
& 73.05 & \textbf{96.94} & -0.94 & \textbf{+56.98} \\

FLTrust
& PFedBA
& 76.08 & 33.34
& 74.57 & 96.49 & -1.50 & +63.15
& 74.86 & \textbf{98.25} & -1.22 & \textbf{+64.91} \\

FLAME
& A3FL
& 75.22 & 13.17
& 74.88 & 93.93 & -0.33 & +80.77
& 74.92 & \textbf{97.92} & -0.30 & \textbf{+84.75} \\

CRFL
& PFedBA
& 72.75 & 22.31
& 68.91 & 93.92 & -3.84 & +71.61
& 68.95 & \textbf{96.86} & -3.80 & \textbf{+74.55} \\

Bucketing
& PFedBA
& 77.33 & 60.09
& 75.84 & 96.08 & -1.49 & +35.99
& 75.49 & \textbf{98.16} & -1.84 & \textbf{+38.07} \\
\bottomrule
\end{tabular}}
\vspace{-0.2cm}
\end{table}

\noindent\textbf{Robustness against defenses.}
We next examine whether ShadowFPT remains effective against aggregation and defense rules that act on uploaded client updates. As shown in Table~\ref{tab:defense_robustness}, ShadowFPT-II achieves the highest ASR under every defense, ranging from 96.63\% to 98.30\%. In contrast, the strongest non-ShadowFPT baseline reaches only 13.17\% to 60.93\% ASR. The gains are especially pronounced under robust and backdoor-specific defenses, including Krum (+73.97 points), LaSA (+76.55 points), FLAME (+84.75 points), and CRFL (+74.55 points).

These results support our claim that update-centered defenses are insufficient for FPT. Although such defenses mitigate abnormal client updates through robust aggregation, filtering, clustering, clipping, or certification, ShadowFPT shifts much of the backdoor effect to the interaction between the adapted trigger and the frozen representation space. ShadowFPT-II improves ASR by 63.61 percentage points on average over Best Base, while incurring only a 1.71-point average clean-accuracy drop. This indicates that its robustness does not primarily come from degrading benign performance.

\begin{figure}[!htbp]
   \includegraphics[width=0.8\linewidth]{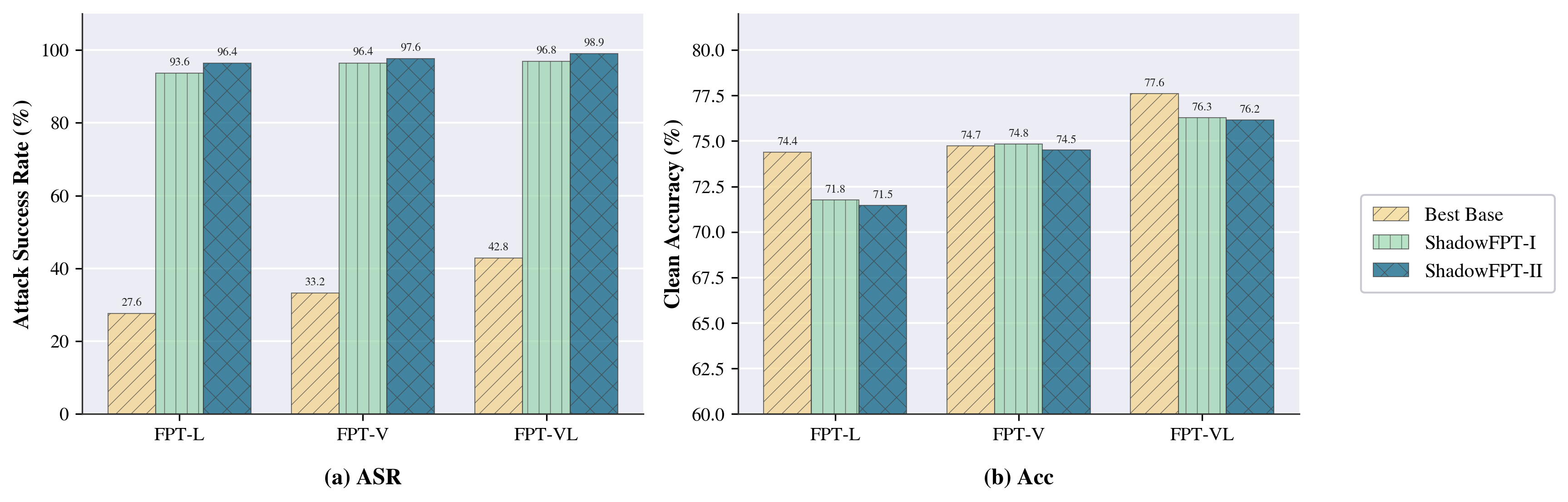}
   \vspace{-0.1cm}
    \caption{
    Transferability of ShadowFPT across prompt modalities. 
    Results are averaged over Caltech101, Oxford Flowers, UCF101, and all aggregation or defense rules. 
    Best Base denotes the strongest non-ShadowFPT baseline under each dataset-setting pair, selected by the highest average ASR before averaging across datasets. 
    ShadowFPT achieves consistently high ASR under FPT-L, FPT-V, and FPT-VL, while largely preserving clean accuracy.
    }
    \label{fig:prompt_modality_transfer}
\end{figure}

\noindent\textbf{Transfer across prompt modalities.}
We further evaluate whether ShadowFPT transfers across textual, visual, and joint vision--language prompt tuning. Figure~\ref{fig:prompt_modality_transfer} reports results averaged over three datasets and all aggregation or defense rules. ShadowFPT-II achieves 96.42\%, 97.63\%, and 98.95\% ASR under FPT-L, FPT-V, and FPT-VL, respectively, substantially outperforming Best Base, which reaches only 27.65\%, 33.20\%, and 42.83\% ASR. ShadowFPT-I shows a similar pattern, with ASR consistently above 93\% across all prompt modalities.

Clean accuracy is largely preserved. Compared with Best Base, ShadowFPT-II incurs average clean-accuracy drops of 2.92, 0.24, and 1.45 percentage points under FPT-L, FPT-V, and FPT-VL, respectively. These results show that ShadowFPT is not specific to a single prompt modality. Instead, it exploits the shared frozen CLIP representation space, which remains common across textual, visual, and joint vision--language prompt tuning.

\begin{figure}[!htbp]
  \centering
  \includegraphics[width=1\linewidth]{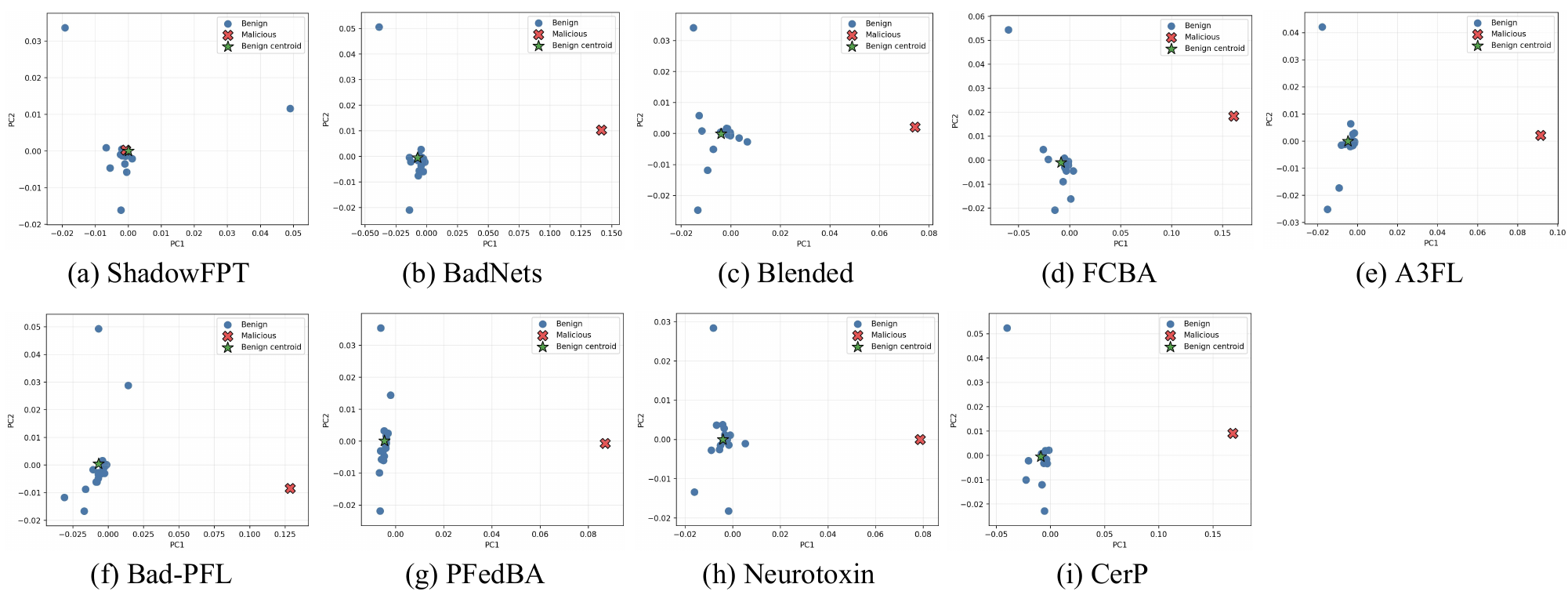}
  \caption{PCA visualization of client prompts under FedAvg. Blue points denote benign clients, the red cross denotes the malicious client, and the green star denotes the benign centroid.}
  \label{fig:pca_prompt}
  \vspace{-0.2cm}
\end{figure}

\noindent\textbf{Prompt-space stealthiness.}
We further examine whether malicious prompt updates can be separated from benign ones. Figure~\ref{fig:pca_prompt} projects client prompts under FedAvg onto the first two principal components. Existing attacks produce malicious prompts that appear as clear outliers from the benign cluster. In contrast, the ShadowFPT prompt lies close to benign clients and nearly overlaps with the benign centroid. This supports our claim that ShadowFPT shifts much of the backdoor effect from uploaded prompt updates to trigger-induced representation steering in the frozen backbone, thereby preserving prompt-space stealthiness.

\subsection{Ablation Study}


\begin{figure}[!htbp]
  \centering
  \begin{minipage}[t]{0.48\linewidth}
    \centering
    \includegraphics[width=\linewidth]{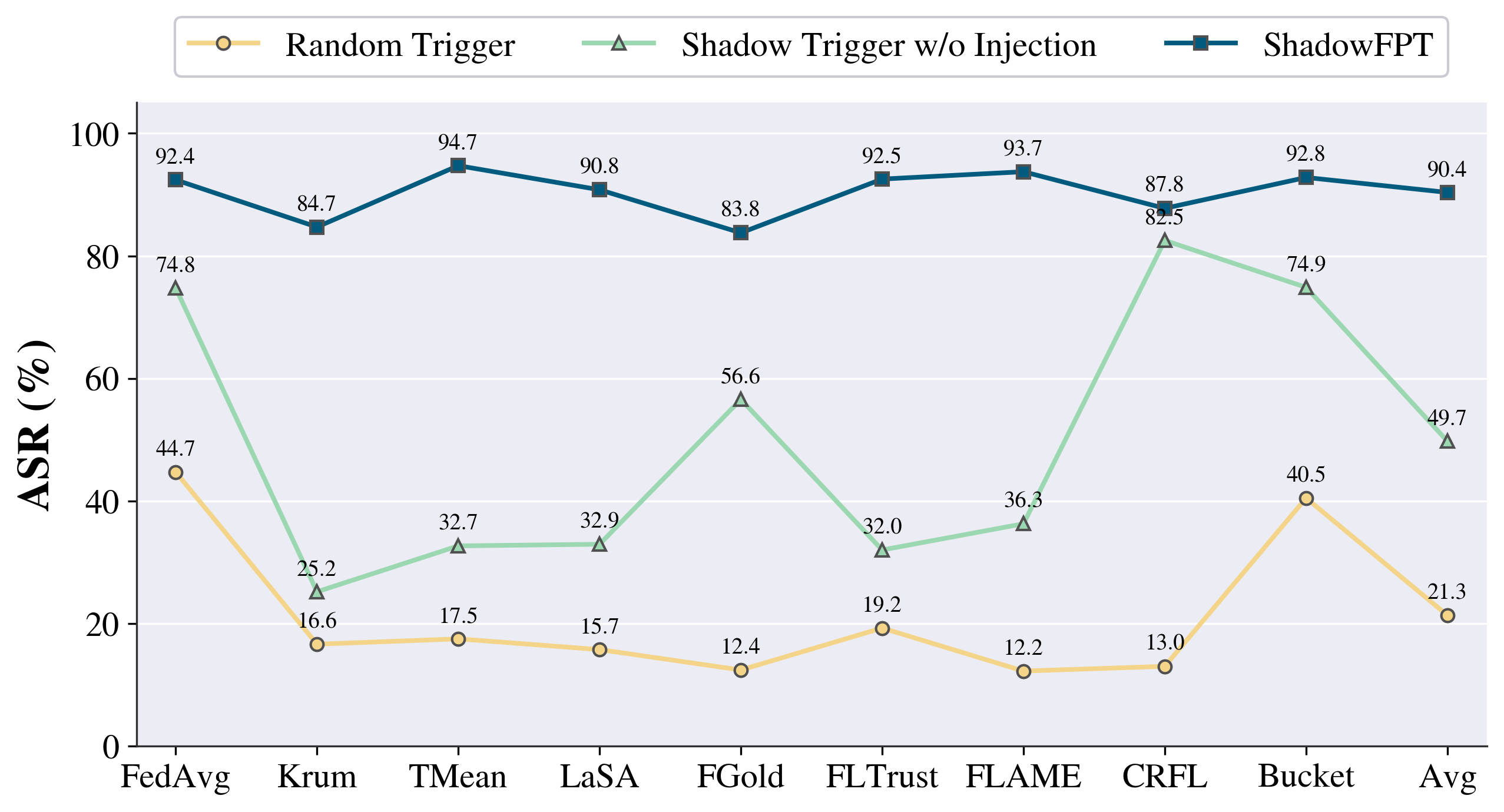}
    \caption{
    Role of Shadow Trigger Pretraining and Backdoor Injection in FPT on Caltech101. Random Trigger uses a randomly initialized trigger; Shadow Trigger w/o Injection uses the pretrained Shadow Trigger without Backdoor Injection in FPT; ShadowFPT denotes the full method.
    }
    \label{fig:aba1}
  \end{minipage}\hfill
  \begin{minipage}[t]{0.48\linewidth}
    \centering
    \includegraphics[width=\linewidth]{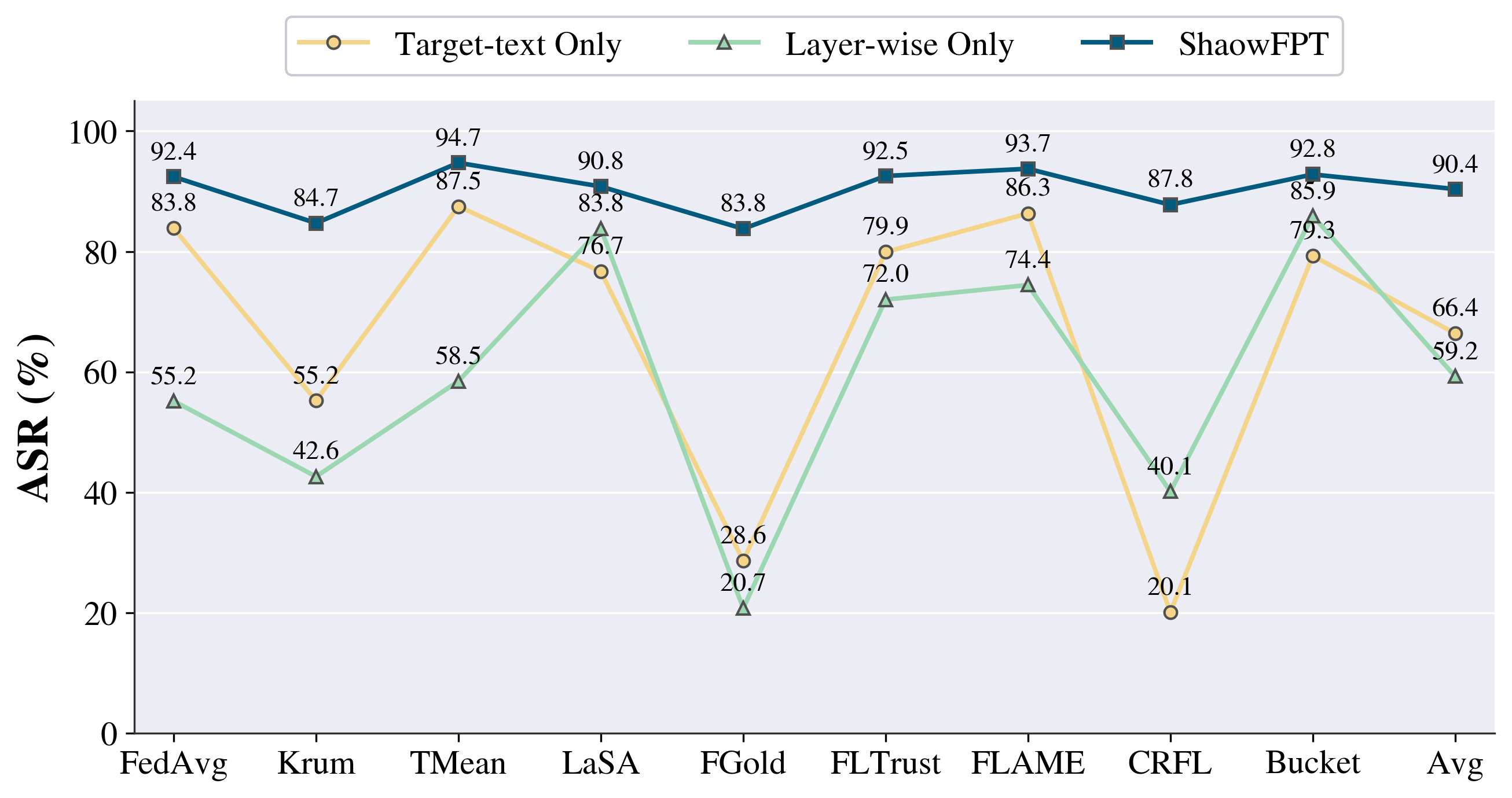}
\caption{
Role of representation-level alignment on Caltech101. ASR is reported under nine aggregation and defense rules. Target-text Only and Layer-wise Only use a single alignment objective, whereas ShadowFPT combines both.
}
    \label{fig:aba2}
  \end{minipage}
\end{figure}

\noindent\textbf{Role of Pretraining and Injection.}
We assess the roles of Shadow Trigger Pretraining and Backdoor Injection in FPT on Caltech101. As shown in Figure~\ref{fig:aba1}, Random Trigger achieves only 21.32\% average ASR, indicating that an arbitrary trigger is insufficient to induce a strong backdoor. Shadow Trigger without Injection increases the average ASR to 49.75\%, suggesting that pretraining already learns target-directed behavior in the frozen CLIP representation space, but remains fragile under defenses such as Krum (25.17\%) and FLTrust (31.99\%). The full ShadowFPT further raises the average ASR to 90.36\%, with large gains under Krum (+59.54) and FLTrust (+60.55). These results show that pretraining provides a transferable initialization, while Backdoor Injection is necessary to convert it into a robust federated backdoor.

\noindent\textbf{Role of Representation-Level Alignment.}
We further examine the two representation-level objectives used in Shadow Trigger Pretraining. As shown in Figure~\ref{fig:aba2}, target-text alignment alone and layer-wise visual alignment alone achieve 66.38\% and 59.25\% average ASR, respectively, indicating that both provide useful target-directed guidance. However, each objective alone remains fragile under defenses, such as CRFL for Target-text Only (20.08\%) and FoolsGold for Layer-wise Only (20.73\%). In contrast, the full ShadowFPT objective reaches 90.36\% average ASR, improving over the two single-objective variants by 23.98 and 31.11 percentage points. These results show that target-text alignment provides semantic target guidance, while layer-wise visual alignment stabilizes the triggered representation trajectory, together producing a more robust Shadow Trigger.
\section{Conclusion}
\label{sec:conclusion}
This paper identifies a frozen-backbone backdoor surface in FPT. Unlike conventional federated backdoor attacks that rely primarily on anomalous uploaded updates, ShadowFPT exploits the shared frozen CLIP representation space and shifts much of the attack burden to trigger-induced representation steering. We propose a two-stage attack in which a Shadow Trigger is first pretrained through target-text and layer-wise visual alignment, and then adapted during FPT while only prompt parameters are uploaded to the server. Across multiple datasets, aggregation and defense rules, and textual, visual, and joint vision--language prompt tuning, ShadowFPT achieves consistently high targeted attack success with limited clean accuracy degradation. Prompt-space analyses further show that its malicious updates remain close to benign prompts, explaining why update-centered defenses are insufficient. Our results suggest that securing FPT requires moving beyond prompt anomaly detection and accounting for the interaction between triggers and frozen representation geometry. More broadly, this work highlights a security risk of FPT: freezing the backbone reduces trainable parameters, but does not eliminate backdoor attack surfaces.

\section*{Acknowledgments}
This work is supported by the National Natural Science Foundation of China (Grant No. 62521004).

\bibliographystyle{plainnat}
\bibliography{ref}

\section{Appendix}\label{app}

\begin{table}[!htbp]
\centering
\caption{
Acc and ASR (\%) on Caltech101, Oxford Flowers, and UCF101 under FPT-L, FPT-V, and FPT-VL. Bold values indicate the best result among all attack methods under each defense. The arrowed values below ShadowFPT-I and ShadowFPT-II report absolute differences from the strongest baseline, where green and red arrows indicate improvement and degradation, respectively.
}
\label{tab:full_results}
\scriptsize
\setlength{\tabcolsep}{2pt}
\resizebox{\linewidth}{!}{
\begin{tabular}{lll*{9}{cc}|cc}
\toprule
\multirow{2}{*}{\rotatebox[origin=c]{45}{Dataset}} & \multirow{2}{*}{\rotatebox[origin=c]{45}{Setting}} & \multirow{2}{*}{\rotatebox[origin=c]{45}{Method}}
& \multicolumn{2}{c}{FedAvg}
& \multicolumn{2}{c}{Krum}
& \multicolumn{2}{c}{TrimmedMean}
& \multicolumn{2}{c}{LaSA}
& \multicolumn{2}{c}{FoolsGold}
& \multicolumn{2}{c}{FLTrust}
& \multicolumn{2}{c}{FLAME}
& \multicolumn{2}{c}{CRFL}
& \multicolumn{2}{c}{Bucketing}
& \multicolumn{2}{|c}{Average} \\
\cmidrule(lr){4-5}
\cmidrule(lr){6-7}
\cmidrule(lr){8-9}
\cmidrule(lr){10-11}
\cmidrule(lr){12-13}
\cmidrule(lr){14-15}
\cmidrule(lr){16-17}
\cmidrule(lr){18-19}
\cmidrule(lr){20-21}
\cmidrule(lr){22-23}
& & & Acc & ASR & Acc & ASR & Acc & ASR & Acc & ASR & Acc & ASR & Acc & ASR & Acc & ASR & Acc & ASR & Acc & ASR & Acc & ASR \\
\midrule
\multirow{36}{*}{\rotatebox[origin=c]{90}{Caltech101}} & \multirow{12}{*}{\rotatebox[origin=c]{90}{FPT-L}}
& BadNets & 91.92 & 2.09 & 89.04 & 0.00 & 93.46 & 0.00 & 92.33 & 0.00 & 76.75 & 0.00 & 90.58 & 0.00 & 89.73 & 0.00 & 75.41 & 0.00 & 92.86 & 0.38 & 88.01 & 0.27 \\
&& Blended & 92.49 & 24.49 & \textbf{89.13} & 0.00 & 93.71 & 0.55 & 93.30 & 0.00 & 75.29 & 0.00 & \textbf{91.44} & 0.00 & 89.73 & 0.00 & 79.83 & 0.00 & \textbf{93.79} & 71.56 & \textbf{88.75} & 10.73 \\
&& FCBA & 92.37 & 1.32 & 88.68 & 0.00 & \textbf{93.50} & 0.00 & 93.10 & 0.00 & 78.78 & 0.00 & 90.46 & 0.00 & 89.73 & 0.00 & 75.37 & 0.00 & 91.76 & 0.43 & 88.19 & 0.19 \\
&& A3FL & 93.34 & 64.58 & 85.03 & 0.17 & 92.45 & 2.14 & 91.68 & 0.08 & 81.17 & 0.00 & 90.38 & 19.35 & 89.00 & 0.34 & 79.92 & 0.00 & 93.46 & 91.64 & 88.49 & 19.81 \\
&& Bad-PFL & \textbf{94.15} & 20.94 & 88.43 & 0.08 & 93.22 & 0.04 & 93.42 & 0.08 & 76.59 & 0.00 & 91.19 & 0.00 & 89.16 & 0.00 & \textbf{79.87} & 0.00 & 92.65 & 8.00 & 88.74 & 3.24 \\
&& PFedBA & 93.46 & 13.10 & 83.85 & 0.08 & 92.61 & 21.58 & 90.62 & 12.46 & 76.59 & 0.00 & 89.77 & 0.56 & 90.18 & 6.55 & 79.51 & 1.93 & 92.94 & 50.79 & 87.73 & 11.89 \\
&& Neurotoxin & 92.98 & 10.03 & 85.43 & 0.00 & 92.49 & 0.00 & 91.68 & 0.00 & 72.45 & 0.00 & 91.11 & 0.00 & 90.18 & 0.00 & 78.45 & 0.00 & 91.07 & 0.00 & 87.32 & 1.11 \\
&& CerP & 87.87 & 10.96 & 89.08 & 0.00 & 92.77 & 0.00 & \textbf{93.95} & 0.00 & 68.92 & 0.00 & 90.54 & 0.00 & 89.61 & 0.00 & 78.10 & 0.00 & 81.74 & 2.74 & 85.84 & 1.52 \\
\rowcolor{darkgray!12}
&& ShadowFPT-I & 91.52 & 90.96 & 83.00 & \textbf{89.29} & 92.29 & 89.20 & 92.49 & 87.49 & \textbf{93.38} & 83.23 & 89.49 & 89.46 & 88.96 & 69.16 & 72.17 & 72.22 & 91.60 & 87.53 & 88.32 & 84.28 \\
&& & \textcolor{red}{(2.63$\downarrow$)}
& \textcolor{green}{(26.38$\uparrow$)}
& \textcolor{red}{(6.13$\downarrow$)}
& \textcolor{green}{(89.12$\uparrow$)}
& \textcolor{red}{(1.42$\downarrow$)}
& \textcolor{green}{(67.62$\uparrow$)}
& \textcolor{red}{(1.46$\downarrow$)}
& \textcolor{green}{(75.03$\uparrow$)}
& \textcolor{green}{(12.21$\uparrow$)}
& \textcolor{green}{(83.23$\uparrow$)}
& \textcolor{red}{(1.95$\downarrow$)}
& \textcolor{green}{(70.11$\uparrow$)}
& \textcolor{red}{(1.22$\downarrow$)}
& \textcolor{green}{(62.61$\uparrow$)}
& \textcolor{red}{(7.75$\downarrow$)}
& \textcolor{green}{(70.29$\uparrow$)}
& \textcolor{red}{(2.19$\downarrow$)}
& \textcolor{red}{(4.11$\downarrow$)}
& \textcolor{red}{(0.43$\downarrow$)}
& \textcolor{green}{(64.47$\uparrow$)} \\

\rowcolor{darkgray!12}
&& ShadowFPT-II & 92.90 & \textbf{92.41} & 83.24 & 84.71 & 92.41 & \textbf{94.73} & 91.19 & \textbf{90.79} & 77.16 & \textbf{83.76} & 91.72 & \textbf{92.54} & \textbf{91.19} & \textbf{93.74} & 76.38 & \textbf{87.76} & 92.86 & \textbf{92.80} & 87.67 & \textbf{90.36} \\

&& & \textcolor{red}{(1.25$\downarrow$)}
& \textcolor{green}{(27.83$\uparrow$)}
& \textcolor{red}{(5.89$\downarrow$)}
& \textcolor{green}{(84.54$\uparrow$)}
& \textcolor{red}{(1.30$\downarrow$)}
& \textcolor{green}{(73.15$\uparrow$)}
& \textcolor{red}{(2.76$\downarrow$)}
& \textcolor{green}{(78.33$\uparrow$)}
& \textcolor{red}{(4.01$\downarrow$)}
& \textcolor{green}{(83.76$\uparrow$)}
& \textcolor{green}{(0.28$\uparrow$)}
& \textcolor{green}{(73.19$\uparrow$)}
& \textcolor{green}{(1.01$\uparrow$)}
& \textcolor{green}{(87.19$\uparrow$)}
& \textcolor{red}{(3.54$\downarrow$)}
& \textcolor{green}{(85.83$\uparrow$)}
& \textcolor{red}{(0.93$\downarrow$)}
& \textcolor{green}{(1.16$\uparrow$)}
& \textcolor{red}{(1.08$\downarrow$)}
& \textcolor{green}{(70.55$\uparrow$)} \\

\cmidrule(lr){2-23}

& \multirow{12}{*}{\rotatebox[origin=c]{90}{FPT-V}}
& BadNets & 93.71 & 4.63 & 89.06 & 0.00 & \textbf{93.46} & 0.00 & 92.33 & 0.00 & 76.75 & 0.00 & 90.58 & 0.00 & 89.73 & 0.00 & 75.41 & 0.00 & 92.86 & 0.28 & 88.21 & 0.55 \\
&& Blended & 92.77 & 10.08 & 92.77 & 0.30 & 93.02 & 0.04 & 92.57 & 0.04 & 91.84 & 1.97 & \textbf{93.54} & 2.22 & 92.41 & 0.10 & 92.61 & 0.32 & 92.25 & 0.13 & 92.64 & 1.69 \\
&& FCBA & 94.03 & 5.04 & 92.34 & 0.00 & 92.94 & 0.00 & 92.61 & 0.00 & 91.48 & 0.98 & 92.17 & 0.00 & 92.45 & 0.00 & \textbf{92.62} & 0.00 & 92.61 & 0.00 & 92.58 & 0.67 \\
&& A3FL & 94.12 & 56.67 & 92.23 & 3.52 & 92.94 & 4.88 & 92.47 & 16.40 & 92.57 & 0.00 & 93.38 & 0.04 & 92.45 & 23.08 & 88.70 & 0.00 & 92.46 & 26.38 & 92.37 & 14.55 \\
&& Bad-PFL & \textbf{94.15} & 20.94 & 88.43 & 0.08 & 93.22 & 0.04 & \textbf{93.42} & 0.08 & 74.92 & 0.00 & 91.19 & 0.00 & 89.16 & 0.00 & 79.87 & 0.00 & 92.65 & 8.00 & 88.56 & 3.24 \\
&& PFedBA & 93.63 & 66.98 & 91.33 & 8.01 & 92.89 & 29.20 & 92.61 & 22.14 & \textbf{94.27} & 71.77 & 93.06 & 60.38 & 92.45 & 0.98 & 90.71 & 20.12 & \textbf{93.59} & 60.99 & \textbf{92.73} & 37.84 \\
&& Neurotoxin & 92.81 & 4.22 & 91.28 & 0.00 & 91.10 & 0.00 & 92.53 & 0.00 & 86.81 & 0.00 & 92.25 & 0.00 & 92.41 & 0.00 & 92.45 & 0.00 & 92.57 & 0.00 & 91.58 & 0.47 \\
&& CerP & 92.41 & 6.50 & \textbf{93.02} & 0.25 & 92.19 & 0.46 & 92.49 & 0.08 & 91.80 & 11.13 & 92.61 & 23.94 & \textbf{92.49} & 0.26 & 92.53 & 0.25 & 92.49 & 0.00 & 92.45 & 4.76 \\
\rowcolor{darkgray!12}
&& ShadowFPT-I & 92.90 & 93.61 & 92.61 & 92.80 & 92.86 & 92.63 & 92.29 & 93.27 & 92.53 & 93.17 & 92.82 & 93.23 & 92.45 & 93.10 & 92.13 & \textbf{92.33} & 92.98 & 92.24 & 92.62 & 92.93 \\
&& & \textcolor{red}{(1.25$\downarrow$)}
& \textcolor{green}{(26.63$\uparrow$)}
& \textcolor{red}{(0.41$\downarrow$)}
& \textcolor{green}{(84.79$\uparrow$)}
& \textcolor{red}{(0.60$\downarrow$)}
& \textcolor{green}{(63.43$\uparrow$)}
& \textcolor{red}{(1.13$\downarrow$)}
& \textcolor{green}{(71.13$\uparrow$)}
& \textcolor{red}{(1.74$\downarrow$)}
& \textcolor{green}{(21.40$\uparrow$)}
& \textcolor{red}{(0.72$\downarrow$)}
& \textcolor{green}{(32.85$\uparrow$)}
& \textcolor{red}{(0.04$\downarrow$)}
& \textcolor{green}{(70.02$\uparrow$)}
& \textcolor{red}{(0.49$\downarrow$)}
& \textcolor{green}{(72.21$\uparrow$)}
& \textcolor{red}{(0.61$\downarrow$)}
& \textcolor{green}{(31.25$\uparrow$)}
& \textcolor{red}{(0.11$\downarrow$)}
& \textcolor{green}{(55.09$\uparrow$)} \\
\rowcolor{darkgray!12}
&& ShadowFPT-II & 92.57 & \textbf{94.64} & 92.62 & \textbf{94.69} & 92.74 & \textbf{94.77} & 92.08 & \textbf{95.37} & 92.94 & \textbf{94.34} & 92.45 & \textbf{94.39} & 92.37 & \textbf{94.52} & 84.41 & 91.33 & 92.94 & \textbf{94.17} & 91.68 & \textbf{94.25} \\
&& & \textcolor{red}{(1.58$\downarrow$)}
& \textcolor{green}{(27.66$\uparrow$)}
& \textcolor{red}{(0.40$\downarrow$)}
& \textcolor{green}{(86.68$\uparrow$)}
& \textcolor{red}{(0.72$\downarrow$)}
& \textcolor{green}{(65.57$\uparrow$)}
& \textcolor{red}{(1.34$\downarrow$)}
& \textcolor{green}{(73.23$\uparrow$)}
& \textcolor{red}{(1.33$\downarrow$)}
& \textcolor{green}{(22.57$\uparrow$)}
& \textcolor{red}{(1.09$\downarrow$)}
& \textcolor{green}{(34.01$\uparrow$)}
& \textcolor{red}{(0.12$\downarrow$)}
& \textcolor{green}{(71.44$\uparrow$)}
& \textcolor{red}{(8.21$\downarrow$)}
& \textcolor{green}{(71.21$\uparrow$)}
& \textcolor{red}{(0.65$\downarrow$)}
& \textcolor{green}{(33.18$\uparrow$)}
& \textcolor{red}{(1.05$\downarrow$)}
& \textcolor{green}{(56.41$\uparrow$)} \\

\cmidrule(lr){2-23}

& \multirow{12}{*}{\rotatebox[origin=c]{90}{FPT-VL}}
& BadNets & 93.02 & 21.02 & 91.60 & 0.00 & 93.26 & 0.08 & 93.43 & 0.13 & 93.54 & 85.61 & 93.31 & 0.04 & 92.49 & 0.04 & 91.76 & 0.00 & 93.43 & 15.76 & 92.87 & 13.63 \\
&& Blended & 93.43 & 77.30 & 91.60 & 0.00 & 93.22 & 1.58 & 93.22 & 48.09 & 93.50 & 77.96 & \textbf{93.38} & 0.81 & 92.49 & 0.00 & \textbf{91.88} & 0.00 & 93.42 & 80.59 & \textbf{92.90} & 31.81 \\
&& FCBA & 91.77 & 4.41 & 91.56 & 0.00 & \textbf{93.38} & 0.04 & 93.43 & 0.17 & 91.19 & 3.43 & 93.30 & 0.04 & 92.49 & 0.04 & 91.76 & 0.00 & \textbf{93.59} & 1.92 & 92.50 & 1.12 \\
&& A3FL & 93.36 & 96.01 & \textbf{91.80} & 0.85 & 93.14 & 0.29 & 93.18 & 93.22 & 92.77 & 80.08 & 93.24 & 37.61 & 92.37 & 45.61 & 91.80 & 0.00 & 93.06 & 95.41 & 92.75 & 49.90 \\
&& Bad-PFL & 93.38 & 70.32 & 91.60 & 33.36 & 93.14 & 62.01 & \textbf{93.51} & 70.32 & 93.02 & 82.35 & 93.31 & 36.74 & 92.61 & 16.15 & 91.80 & 2.35 & 92.94 & 81.75 & 92.83 & 50.59 \\
&& PFedBA & 93.38 & 70.32 & 91.78 & 33.36 & 93.14 & 62.01 & 93.42 & 77.47 & 93.02 & 82.35 & 93.30 & 36.74 & 92.61 & 16.14 & 91.80 & 2.35 & 92.94 & 81.75 & 92.82 & 51.39 \\
&& Neurotoxin & 93.14 & 1.19 & 91.72 & 0.00 & 93.26 & 0.08 & 93.38 & 0.08 & 90.99 & 2.69 & 93.22 & 0.08 & \textbf{92.65} & 0.04 & 91.76 & 0.00 & 93.18 & 0.04 & 92.59 & 0.47 \\
&& CerP & 80.00 & 30.70 & 91.76 & 0.00 & 93.26 & 0.08 & 85.27 & 23.85 & 73.99 & 27.64 & 93.34 & 0.08 & 92.61 & 0.02 & 91.72 & 0.00 & 65.07 & 42.86 & 85.23 & 13.91 \\
\rowcolor{darkgray!12}
&& ShadowFPT-I & \textbf{93.73} & 93.06 & 90.79 & 92.51 & 92.21 & 92.50 & 92.49 & 93.36 & \textbf{93.71} & 96.53 & 93.18 & 94.56 & 91.52 & 91.86 & 91.15 & 91.43 & 92.77 & 92.93 & 92.39 & 93.19 \\
&& &\textcolor{green}{(0.30$\uparrow$)}
& \textcolor{red}{(2.95$\downarrow$)}
& \textcolor{red}{(1.01$\downarrow$)}
& \textcolor{green}{(59.15$\uparrow$)}
& \textcolor{red}{(1.17$\downarrow$)}
& \textcolor{green}{(30.49$\uparrow$)}
& \textcolor{red}{(1.02$\downarrow$)}
& \textcolor{green}{(0.14$\uparrow$)}
& \textcolor{green}{(0.17$\uparrow$)}
& \textcolor{green}{(10.92$\uparrow$)}
& \textcolor{red}{(0.20$\downarrow$)}
& \textcolor{green}{(56.95$\uparrow$)}
& \textcolor{red}{(1.13$\downarrow$)}
& \textcolor{green}{(46.25$\uparrow$)}
& \textcolor{red}{(0.73$\downarrow$)}
& \textcolor{green}{(89.08$\uparrow$)}
& \textcolor{red}{(0.82$\downarrow$)}
& \textcolor{red}{(2.48$\downarrow$)}
& \textcolor{red}{(0.51$\downarrow$)}
& \textcolor{green}{(41.80$\uparrow$)} \\
\rowcolor{darkgray!12}
&& ShadowFPT-II & 92.73 & \textbf{97.73} & 90.79 & \textbf{97.34} & 92.21 & \textbf{98.02} & 92.53 & \textbf{98.24} & 93.67 & \textbf{98.75} & 93.18 & \textbf{98.41} & 91.52 & \textbf{98.02} & 91.15 & \textbf{97.55} & 92.77 & \textbf{98.24} & 92.28 & \textbf{98.03} \\
&& & \textcolor{red}{(0.70$\downarrow$)}
& \textcolor{green}{(1.72$\uparrow$)}
& \textcolor{red}{(1.01$\downarrow$)}
& \textcolor{green}{(63.98$\uparrow$)}
& \textcolor{red}{(1.17$\downarrow$)}
& \textcolor{green}{(36.01$\uparrow$)}
& \textcolor{red}{(0.98$\downarrow$)}
& \textcolor{green}{(5.02$\uparrow$)}
& \textcolor{green}{(0.13$\uparrow$)}
& \textcolor{green}{(13.14$\uparrow$)}
& \textcolor{red}{(0.20$\downarrow$)}
& \textcolor{green}{(60.80$\uparrow$)}
& \textcolor{red}{(1.13$\downarrow$)}
& \textcolor{green}{(52.41$\uparrow$)}
& \textcolor{red}{(0.73$\downarrow$)}
& \textcolor{green}{(95.20$\uparrow$)}
& \textcolor{red}{(0.82$\downarrow$)}
& \textcolor{green}{(2.83$\uparrow$)}
& \textcolor{red}{(0.62$\downarrow$)}
& \textcolor{green}{(46.64$\uparrow$)} \\

\midrule

\multirow{36}{*}{\rotatebox[origin=c]{90}{Oxford Flowers}} & \multirow{12}{*}{\rotatebox[origin=c]{90}{FPT-L}}
& BadNets & 67.88 & 6.89 & 64.39 & 0.28 & 68.69 & 1.22 & 67.47 & 3.05 & 68.21 & 14.15 & 64.31 & 0.24 & 67.47 & 0.37 & 44.25 & 0.41 & 65.44 & 5.54 & 64.23 & 3.57 \\
&& Blended & 69.99 & 27.91 & 64.39 & 0.32 & 69.14 & 5.75 & 67.07 & 25.58 & 69.26 & 25.18 & 65.97 & 6.28 & 66.74 & 0.04 & 44.33 & 0.53 & 64.72 & 30.39 & 64.62 & 13.55 \\
&& FCBA & 68.37 & 6.85 & 64.39 & 0.36 & 68.33 & 1.14 & 69.87 & 2.03 & 65.93 & 6.85 & 66.82 & 0.69 & 66.05 & 0.53 & 43.11 & 0.40 & 66.13 & 3.42 & 64.33 & 2.47 \\
&& A3FL & 70.76 & 97.87 & 65.00 & 0.00 & 67.07 & 93.96 & 70.11 & 17.78 & 71.98 & 29.49 & 68.16 & 81.19 & 64.63 & 18.48 & 51.44 & 0.00 & \textbf{70.52} & 3.83 & 66.63 & 38.07 \\
&& Bad-PFL & 69.71 & 7.71 & 62.72 & 0.40 & 68.29 & 4.93 & 69.06 & 6.52 & \textbf{76.20} & 10.15 & 68.89 & 4.03 & 66.34 & 2.40 & 49.24 & 0.97 & 69.63 & 10.24 & 66.68 & 5.26 \\
&& PFedBA & 70.92 & 33.51 & \textbf{69.38} & 24.12 & \textbf{72.71} & 26.67 & 71.33 & 4.20 & 68.41 & 32.46 & 67.52 & 30.79 & \textbf{69.67} & 14.64 & \textbf{68.77} & 35.07 & 70.28 & 29.67 & \textbf{69.89} & 25.68 \\
&& Neurotoxin & \textbf{71.78} & 3.34 & 66.66 & 0.34 & 68.53 & 0.85 & 68.73 & 1.71 & 66.86 & 2.12 & \textbf{69.63} & 0.85 & 67.07 & 0.57 & 48.55 & 0.16 & 67.68 & 1.06 & 66.17 & 1.22 \\
&& CerP & 69.26 & 5.26 & 63.29 & 0.16 & 68.65 & 0.85 & \textbf{71.73} & 1.06 & 65.81 & 7.62 & 66.82 & 1.67 & 67.35 & 0.57 & 51.92 & 0.24 & 63.82 & 13.83 & 65.41 & 3.47 \\
\rowcolor{darkgray!12}
&& ShadowFPT-I & 62.03 & 96.89 & 54.42 & 94.90 & 62.07 & 97.12 & 65.21 & 96.73 & 62.89 & 98.0 & 63.05 & 97.22 & 63.94 & 97.31 & 49.20 & 96.16 & 65.52 & 96.36 & 60.93 & 96.74 \\
&& & \textcolor{red}{(9.75$\downarrow$)}
& \textcolor{red}{(0.98$\downarrow$)}
& \textcolor{red}{(14.96$\downarrow$)}
& \textcolor{green}{(70.78$\uparrow$)}
& \textcolor{red}{(10.64$\downarrow$)}
& \textcolor{green}{(3.16$\uparrow$)}
& \textcolor{red}{(6.52$\downarrow$)}
& \textcolor{green}{(71.15$\uparrow$)}
& \textcolor{red}{(13.31$\downarrow$)}
& \textcolor{green}{(65.54$\uparrow$)}
& \textcolor{red}{(6.58$\downarrow$)}
& \textcolor{green}{(16.03$\uparrow$)}
& \textcolor{red}{(5.73$\downarrow$)}
& \textcolor{green}{(78.83$\uparrow$)}
& \textcolor{red}{(19.57$\downarrow$)}
& \textcolor{green}{(61.09$\uparrow$)}
& \textcolor{red}{(5.00$\downarrow$)}
& \textcolor{green}{(65.97$\uparrow$)}
& \textcolor{red}{(8.96$\downarrow$)}
& \textcolor{green}{(58.67$\uparrow$)} \\
\rowcolor{darkgray!12}
&& ShadowFPT-II & 63.12 & \textbf{100.0} & 58.66 & \textbf{98.33} & 61.87 & \textbf{98.74} & 65.18 & \textbf{98.54} & 58.66 & \textbf{99.10} & 63.13 & \textbf{99.81} & 63.42 & \textbf{98.22} & 53.33 & \textbf{97.91} & 60.29 & \textbf{99.76} & 60.85 & \textbf{98.93} \\
&& & \textcolor{red}{(8.66$\downarrow$)}
& \textcolor{green}{(2.13$\uparrow$)}
& \textcolor{red}{(10.72$\downarrow$)}
& \textcolor{green}{(74.21$\uparrow$)}
& \textcolor{red}{(10.84$\downarrow$)}
& \textcolor{green}{(4.78$\uparrow$)}
& \textcolor{red}{(6.55$\downarrow$)}
& \textcolor{green}{(72.96$\uparrow$)}
& \textcolor{red}{(17.54$\downarrow$)}
& \textcolor{green}{(66.64$\uparrow$)}
& \textcolor{red}{(6.50$\downarrow$)}
& \textcolor{green}{(18.62$\uparrow$)}
& \textcolor{red}{(6.25$\downarrow$)}
& \textcolor{green}{(79.74$\uparrow$)}
& \textcolor{red}{(15.44$\downarrow$)}
& \textcolor{green}{(62.84$\uparrow$)}
& \textcolor{red}{(10.23$\downarrow$)}
& \textcolor{green}{(69.37$\uparrow$)}
& \textcolor{red}{(9.04$\downarrow$)}
& \textcolor{green}{(60.86$\uparrow$)} \\
\cmidrule(lr){2-23}

& \multirow{12}{*}{\rotatebox[origin=c]{90}{FPT-V}}
& BadNets & 66.13 & 4.21 & 66.31 & 0.04 & 67.11 & 0.45 & 66.46 & 0.03 & 44.05 & 2.32 & \textbf{69.87} & 0.89 & 66.71 & 0.04 & \textbf{69.28} & 1.21 & 64.88 & 0.97 & 64.53 & 1.13 \\
&& Blended & 66.34 & 26.39 & 66.22 & 0.04 & 68.49 & 3.46 & 66.71 & 0.12 & 48.55 & 0.21 & 68.45 & 2.21 & 66.71 & 0.21 & 68.32 & 0.33 & 64.67 & 11.46 & 64.94 & 4.94 \\
&& FCBA & 67.19 & 3.22 & 66.34 & 0.20 & 66.99 & 0.93 & \textbf{66.74} & 0.16 & 43.44 & 1.83 & 65.21 & 2.36 & 66.62 & 0.12 & 66.21 & 2.33 & 64.88 & 1.87 & 63.74 & 1.45 \\
&& A3FL & 66.91 & 20.27 & 66.82 & 0.89 & 68.29 & 0.00 & 66.73 & 0.00 & 58.79 & 16.40 & 69.06 & 0.00 & 66.71 & 10.85 & 63.09 & 2.65 & 64.92 & 16.55 & 65.70 & 7.51 \\
&& Bad-PFL & 66.54 & 19.09 & 66.91 & 4.85 & \textbf{69.18} & 12.23 & 66.62 & 4.36 & 48.31 & 5.58 & 66.26 & 2.57 & 66.67 & 4.41 & 64.59 & 0.45 & 66.58 & 12.89 & 64.63 & 7.38 \\
&& PFedBA & \textbf{67.61} & 47.98 & 66.74 & 29.95 & 67.92 & 33.31 & 66.70 & 25.49 & 58.67 & 34.96 & 67.11 & 38.32 & 66.94 & 9.18 & 64.27 & 7.46 & 65.57 & 45.32 & 65.73 & 30.22 \\
&& Neurotoxin & 66.95 & 0.97 & 66.62 & 0.04 & 68.45 & 0.36 & 66.71 & 0.04 & 62.93 & 0.57 & 67.80 & 0.24 & 66.71 & 0.04 & 66.45 & 0.42 & 64.55 & 0.69 & 66.35 & 0.37 \\
&& CerP & 67.43 & 21.21 & 66.22 & 1.12 & 68.04 & 5.10 & 66.71 & 7.29 & 62.22 & 2.61 & 67.92 & 4.27 & 66.67 & 15.71 & 65.91 & 4.22 & 66.32 & 6.97 & 66.38 & 7.61 \\
\rowcolor{darkgray!12}
&& ShadowFPT-I & 67.23 & 98.28 & 66.95 & 98.0 & 66.51 & 98.08 & 66.46 & 98.08 & \textbf{67.27} & 98.73 & 63.58 & 99.97 & \textbf{66.95} & 98.20 & 67.03 & 97.59 & 67.39 & 98.41 & 66.60 & 98.37 \\
&& & \textcolor{red}{(0.38$\downarrow$)}
& \textcolor{green}{(50.30$\uparrow$)}
& \textcolor{green}{(0.04$\uparrow$)}
& \textcolor{green}{(68.05$\uparrow$)}
& \textcolor{red}{(2.67$\downarrow$)}
& \textcolor{green}{(64.77$\uparrow$)}
& \textcolor{red}{(0.28$\downarrow$)}
& \textcolor{green}{(72.59$\uparrow$)}
& \textcolor{green}{(4.34$\uparrow$)}
& \textcolor{green}{(63.77$\uparrow$)}
& \textcolor{red}{(6.29$\downarrow$)}
& \textcolor{green}{(61.65$\uparrow$)}
& \textcolor{green}{(0.01$\uparrow$)}
& \textcolor{green}{(82.49$\uparrow$)}
& \textcolor{red}{(2.25$\downarrow$)}
& \textcolor{green}{(90.13$\uparrow$)}
& \textcolor{green}{(0.81$\uparrow$)}
& \textcolor{green}{(53.09$\uparrow$)}
& \textcolor{green}{(0.22$\uparrow$)}
& \textcolor{green}{(68.15$\uparrow$)} \\

\rowcolor{darkgray!12}
&& ShadowFPT-II & 66.86 & \textbf{99.96} & \textbf{67.15} & \textbf{98.21} & 67.13 & \textbf{98.12} & 66.67 & \textbf{99.24} & 66.42 & \textbf{99.44} & 63.94 & \textbf{100.0} & 66.89 & \textbf{98.46} & 66.99 & \textbf{98.12} & \textbf{67.43} & \textbf{98.91} & \textbf{66.61} & \textbf{98.94} \\
&& & \textcolor{red}{(0.75$\downarrow$)}
& \textcolor{green}{(51.98$\uparrow$)}
& \textcolor{green}{(0.24$\uparrow$)}
& \textcolor{green}{(68.26$\uparrow$)}
& \textcolor{red}{(2.05$\downarrow$)}
& \textcolor{green}{(64.81$\uparrow$)}
& \textcolor{red}{(0.07$\downarrow$)}
& \textcolor{green}{(73.75$\uparrow$)}
& \textcolor{green}{(3.49$\uparrow$)}
& \textcolor{green}{(64.48$\uparrow$)}
& \textcolor{red}{(5.93$\downarrow$)}
& \textcolor{green}{(61.68$\uparrow$)}
& \textcolor{red}{(0.05$\downarrow$)}
& \textcolor{green}{(82.75$\uparrow$)}
& \textcolor{red}{(2.29$\downarrow$)}
& \textcolor{green}{(90.66$\uparrow$)}
& \textcolor{green}{(0.85$\uparrow$)}
& \textcolor{green}{(53.59$\uparrow$)}
& \textcolor{green}{(0.23$\uparrow$)}
& \textcolor{green}{(68.72$\uparrow$)} \\
\cmidrule(lr){2-23}

& \multirow{12}{*}{\rotatebox[origin=c]{90}{FPT-VL}}
& BadNets & 70.44 & 4.08 & 69.02 & 0.12 & 71.86 & 0.44 & 71.33 & 0.97 & \textbf{72.95} & 3.53 & 68.69 & 0.73 & 70.41 & 0.21 & 68.77 & 0.41 & 67.76 & 1.91 & 70.14 & 1.38 \\
&& Blended & \textbf{72.21} & 26.81 & 68.98 & 1.50 & 71.78 & 5.14 & 71.57 & 18.64 & 70.97 & 20.43 & 67.76 & 25.88 & \textbf{71.41} & 0.93 & 68.49 & 2.20 & 68.37 & 17.74 & 70.17 & 13.25 \\
&& FCBA & 69.83 & 3.79 & 69.02 & 0.28 & 72.02 & 0.61 & 71.78 & 1.01 & 67.60 & 3.63 & 69.02 & 0.48 & 70.23 & 0.32 & 68.73 & 0.57 & 67.43 & 1.71 & 69.52 & 1.38 \\
&& A3FL & 70.80 & 88.98 & 69.32 & 12.97 & \textbf{72.91} & 3.31 & 70.88 & 6.97 & 68.08 & 73.27 & \textbf{69.79} & 30.64 & 68.89 & 19.46 & 68.49 & 22.19 & 71.74 & 35.78 & 70.10 & 32.62 \\
&& Bad-PFL & 71.33 & 8.48 & 68.62 & 4.81 & 71.98 & 3.83 & \textbf{72.06} & 3.02 & 72.18 & 12.52 & 68.77 & 6.40 & 70.07 & 2.36 & 68.74 & 3.75 & \textbf{72.75} & 6.69 & \textbf{70.72} & 5.76 \\
&& PFedBA & 70.92 & 73.51 & 69.38 & 64.12 & 72.71 & 56.67 & 71.33 & 4.20 & 68.41 & 62.46 & 67.51 & 70.97 & 69.67 & 14.64 & 68.77 & 75.07 & 70.28 & 69.76 & 69.89 & 54.60 \\
&& Neurotoxin & 69.26 & 0.82 & \textbf{70.56} & 0.08 & 72.22 & 0.36 & 70.44 & 0.08 & 69.95 & 3.46 & 68.65 & 0.61 & 69.18 & 0.62 & 68.89 & 0.28 & 68.81 & 0.65 & 69.77 & 0.77 \\
&& CerP & 71.53 & 5.58 & 69.22 & 0.77 & 71.53 & 1.55 & 71.57 & 1.26 & 69.51 & 4.20 & 69.14 & 2.12 & 70.40 & 0.53 & \textbf{68.94} & 0.82 & 67.27 & 5.34 & 69.90 & 2.46 \\
\rowcolor{darkgray!12}
&& ShadowFPT-I & 68.77 & 98.26 & 66.26 & 98.69 & 68.94 & 98.81 & 69.11 & 98.85 & 70.03 & 94.36 & 69.31 & 97.38 & 68.57 & \textbf{98.85} & 66.01 & 99.10 & 69.71 & 98.36 & 68.52 & 98.07 \\
&& & \textcolor{red}{(3.44$\downarrow$)}
& \textcolor{green}{(9.28$\uparrow$)}
& \textcolor{red}{(4.30$\downarrow$)}
& \textcolor{green}{(34.57$\uparrow$)}
& \textcolor{red}{(3.97$\downarrow$)}
& \textcolor{green}{(42.14$\uparrow$)}
& \textcolor{red}{(2.95$\downarrow$)}
& \textcolor{green}{(80.21$\uparrow$)}
& \textcolor{red}{(2.92$\downarrow$)}
& \textcolor{green}{(21.09$\uparrow$)}
& \textcolor{red}{(0.48$\downarrow$)}
& \textcolor{green}{(26.41$\uparrow$)}
& \textcolor{red}{(2.84$\downarrow$)}
& \textcolor{green}{(79.39$\uparrow$)}
& \textcolor{red}{(2.93$\downarrow$)}
& \textcolor{green}{(24.03$\uparrow$)}
& \textcolor{red}{(3.04$\downarrow$)}
& \textcolor{green}{(28.60$\uparrow$)}
& \textcolor{red}{(2.20$\downarrow$)}
& \textcolor{green}{(43.47$\uparrow$)} \\
\rowcolor{darkgray!12}
&& ShadowFPT-II & 69.06 & \textbf{100.0} & 66.31 & \textbf{99.44} & 68.82 & \textbf{98.92} & 70.07 & \textbf{99.31} & 67.72 & \textbf{97.54} & 69.18 & \textbf{99.68} & 68.37 & 98.47 & 65.89 & \textbf{99.29} & 69.58 & \textbf{99.93} & 68.33 & \textbf{99.18} \\
&& & \textcolor{red}{(3.15$\downarrow$)}
& \textcolor{green}{(11.02$\uparrow$)}
& \textcolor{red}{(4.25$\downarrow$)}
& \textcolor{green}{(35.32$\uparrow$)}
& \textcolor{red}{(4.09$\downarrow$)}
& \textcolor{green}{(42.25$\uparrow$)}
& \textcolor{red}{(1.99$\downarrow$)}
& \textcolor{green}{(80.67$\uparrow$)}
& \textcolor{red}{(5.23$\downarrow$)}
& \textcolor{green}{(24.27$\uparrow$)}
& \textcolor{red}{(0.61$\downarrow$)}
& \textcolor{green}{(28.71$\uparrow$)}
& \textcolor{red}{(3.04$\downarrow$)}
& \textcolor{green}{(79.01$\uparrow$)}
& \textcolor{red}{(3.05$\downarrow$)}
& \textcolor{green}{(24.22$\uparrow$)}
& \textcolor{red}{(3.17$\downarrow$)}
& \textcolor{green}{(30.17$\uparrow$)}
& \textcolor{red}{(2.39$\downarrow$)}
& \textcolor{green}{(44.58$\uparrow$)} \\
\midrule
\multirow{36}{*}{\rotatebox[origin=c]{90}{UCF101}} & \multirow{12}{*}{\rotatebox[origin=c]{90}{FPT-L}}
& BadNets & 69.62 & 0.75 & 64.82 & 0.02 & 71.13 & 0.03 & 69.94 & 0.00 & 69.92 & 0.51 & 70.71 & 0.16 & 70.28 & 0.03 & \textbf{62.01} & 0.29 & 69.60 & 0.24 & 68.67 & 0.23 \\
&& Blended & 70.23 & 52.42 & \textbf{64.84} & 0.05 & \textbf{72.19} & 0.64 & 70.36 & 0.26 & \textbf{70.55} & 35.07 & 70.10 & 10.56 & 70.23 & 0.14 & 61.51 & 0.43 & 69.83 & 33.41 & \textbf{68.87} & 14.78 \\
&& FCBA & 70.05 & 0.90 & 64.83 & 0.03 & 71.68 & 0.13 & 70.49 & 0.08 & 69.60 & 0.13 & 69.09 & 0.41 & \textbf{70.34} & 0.02 & 60.61 & 0.37 & 70.44 & 0.18 & 68.57 & 0.25 \\
&& A3FL & 71.13 & 49.87 & 60.03 & 2.51 & 70.36 & 3.63 & 70.26 & 1.42 & 69.78 & 9.36 & \textbf{71.74} & 29.17 & 68.86 & 0.51 & 60.66 & 0.77 & \textbf{72.11} & 14.06 & 68.33 & 12.37 \\
&& Bad-PFL & 70.76 & 0.72 & 54.24 & 0.51 & 71.66 & 0.98 & \textbf{71.23} & 0.55 & 70.33 & 0.67 & 71.45 & 0.51 & 68.51 & 0.41 & 61.69 & 0.51 & 69.71 & 1.04 & 67.73 & 0.66 \\
&& PFedBA & 71.47 & 31.55 & 58.57 & 2.19 & 71.05 & 15.03 & 71.02 & 26.77 & 70.39 & 19.06 & 71.50 & 5.69 & 67.77 & 29.23 & 59.43 & 11.55 & 70.97 & 84.51 & 68.02 & 25.06 \\
&& Neurotoxin & \textbf{72.11} & 0.27 & 59.55 & 0.03 & 70.92 & 0.24 & 70.76 & 0.16 & 68.59 & 0.18 & 71.18 & 0.11 & \textbf{70.34} & 0.05 & 59.43 & 0.43 & 71.29 & 0.05 & 68.24 & 0.17 \\
&& CerP & 68.14 & 4.33 & 59.92 & 0.13 & 71.50 & 0.32 & 70.79 & 0.11 & 57.04 & 48.22 & 69.46 & 0.56 & 69.60 & 0.08 & 61.35 & 0.37 & 68.27 & 1.25 & 66.23 & 6.15 \\
\rowcolor{darkgray!12}
&& ShadowFPT-I & 69.07 & 99.86 & 61.96 & 99.58 & 69.89 & 99.78 & 69.49 & \textbf{100.0} & 68.54 & \textbf{100.0} & 66.98 & \textbf{100.0} & 67.83 & 99.23 & 53.39 & 99.81 & 67.27 & \textbf{100.0} & 66.05 & \textbf{99.81} \\
&& & \textcolor{red}{(3.04$\downarrow$)}
& \textcolor{green}{(47.44$\uparrow$)}
& \textcolor{red}{(2.88$\downarrow$)}
& \textcolor{green}{(97.07$\uparrow$)}
& \textcolor{red}{(2.30$\downarrow$)}
& \textcolor{green}{(84.75$\uparrow$)}
& \textcolor{red}{(1.74$\downarrow$)}
& \textcolor{green}{(73.23$\uparrow$)}
& \textcolor{red}{(2.01$\downarrow$)}
& \textcolor{green}{(51.78$\uparrow$)}
& \textcolor{red}{(4.76$\downarrow$)}
& \textcolor{green}{(70.83$\uparrow$)}
& \textcolor{red}{(2.51$\downarrow$)}
& \textcolor{green}{(70.00$\uparrow$)}
& \textcolor{red}{(8.62$\downarrow$)}
& \textcolor{green}{(88.26$\uparrow$)}
& \textcolor{red}{(4.84$\downarrow$)}
& \textcolor{green}{(15.49$\uparrow$)}
& \textcolor{red}{(2.82$\downarrow$)}
& \textcolor{green}{(74.75$\uparrow$)} \\
\rowcolor{darkgray!12}
&& ShadowFPT-II & 68.78 & \textbf{99.97} & 61.96 & \textbf{99.97} & 70.10 & \textbf{99.97} & 69.69 & 99.97 & 66.05 & \textbf{100.0} & 67.45 & 99.97 & 66.82 & \textbf{99.97} & 53.37 & \textbf{99.98} & 68.59 & \textbf{100.0} & 65.87 & 99.98 \\
&& & \textcolor{red}{(3.33$\downarrow$)}
& \textcolor{green}{(47.55$\uparrow$)}
& \textcolor{red}{(2.88$\downarrow$)}
& \textcolor{green}{(97.46$\uparrow$)}
& \textcolor{red}{(2.09$\downarrow$)}
& \textcolor{green}{(84.94$\uparrow$)}
& \textcolor{red}{(1.54$\downarrow$)}
& \textcolor{green}{(73.20$\uparrow$)}
& \textcolor{red}{(4.50$\downarrow$)}
& \textcolor{green}{(51.78$\uparrow$)}
& \textcolor{red}{(4.29$\downarrow$)}
& \textcolor{green}{(70.80$\uparrow$)}
& \textcolor{red}{(3.52$\downarrow$)}
& \textcolor{green}{(70.74$\uparrow$)}
& \textcolor{red}{(8.64$\downarrow$)}
& \textcolor{green}{(88.43$\uparrow$)}
& \textcolor{red}{(3.52$\downarrow$)}
& \textcolor{green}{(15.49$\uparrow$)}
& \textcolor{red}{(3.00$\downarrow$)}
& \textcolor{green}{(74.92$\uparrow$)} \\
\cmidrule(lr){2-23}

& \multirow{12}{*}{\rotatebox[origin=c]{90}{FPT-V}}
& BadNets & 67.96 & 1.97 & 64.68 & 0.27 & 67.59 & 0.05 & 65.45 & 0.32 & 65.92 & 1.38 & 65.68 & 0.51 & 65.45 & 0.32 & 64.63 & 0.16 & 67.59 & 2.64 & 66.11 & 0.85 \\
&& Blended & 68.09 & 36.26 & 64.48 & 0.49 & 67.77 & 5.61 & 65.45 & 0.61 & \textbf{66.41} & 36.10 & \textbf{68.67} & 21.18 & 65.45 & 0.62 & 64.81 & 0.11 & 64.78 & 12.67 & 66.21 & 12.63 \\
&& FCBA & 67.93 & 0.62 & 64.68 & 0.26 & 67.72 & 0.13 & 65.39 & 0.27 & 64.84 & 0.45 & 67.48 & 0.24 & 65.45 & 0.26 & 64.76 & 0.13 & 67.56 & 0.45 & 66.20 & 0.31 \\
&& A3FL & \textbf{69.69} & 40.24 & \textbf{65.55} & 38.40 & 68.09 & 0.00 & 65.45 & 0.08 & 66.03 & 0.00 & 66.05 & 0.00 & 65.45 & 0.11 & 64.97 & 7.14 & 67.12 & 3.74 & \textbf{66.49} & 9.97 \\
&& Bad-PFL & 68.12 & 1.25 & 65.21 & 0.26 & 68.12 & 0.29 & \textbf{65.47} & 0.29 & 64.71 & 0.53 & 66.82 & 1.12 & 65.23 & 0.32 & 65.00 & 0.13 & \textbf{67.77} & 0.29 & 66.27 & 0.50 \\
&& PFedBA & 67.31 & 39.26 & 64.97 & 28.72 & 67.41 & 20.48 & 65.45 & 0.35 & 64.36 & 28.68 & 64.60 & 32.44 & \textbf{65.47} & 0.72 & 65.02 & 39.95 & 67.24 & 93.39 & 65.76 & 31.55 \\
&& Neurotoxin & 67.77 & 0.05 & 64.68 & 0.27 & \textbf{68.31} & 0.05 & 65.37 & 0.32 & 66.11 & 0.18 & 65.92 & 0.11 & 65.39 & 0.32 & \textbf{65.08} & 0.22 & 67.14 & 0.13 & 66.20 & 0.18 \\
&& CerP & 67.75 & 0.93 & 65.11 & 0.27 & 67.85 & 0.43 & 65.43 & 0.53 & 60.95 & 0.99 & 66.51 & 61.05 & 65.45 & 1.79 & 64.76 & 1.23 & 65.95 & 8.05 & 65.53 & 8.36 \\
\rowcolor{darkgray!12}
&& ShadowFPT-I & 65.21 & 98.73 & 65.53 & 97.56 & 65.29 & 97.28 & 64.89 & 98.02 & 65.55 & 98.69 & 65.53 & 97.62 & 65.32 & 98.35 & 64.78 & 97.01 & 65.46 & 99.08 & 65.28 & 98.04 \\
&& & \textcolor{red}{(4.48$\downarrow$)}
& \textcolor{green}{(58.49$\uparrow$)}
& \textcolor{red}{(0.02$\downarrow$)}
& \textcolor{green}{(59.16$\uparrow$)}
& \textcolor{red}{(3.02$\downarrow$)}
& \textcolor{green}{(76.80$\uparrow$)}
& \textcolor{red}{(0.58$\downarrow$)}
& \textcolor{green}{(97.41$\uparrow$)}
& \textcolor{red}{(0.86$\downarrow$)}
& \textcolor{green}{(62.59$\uparrow$)}
& \textcolor{red}{(3.14$\downarrow$)}
& \textcolor{green}{(36.57$\uparrow$)}
& \textcolor{red}{(0.15$\downarrow$)}
& \textcolor{green}{(96.56$\uparrow$)}
& \textcolor{red}{(0.30$\downarrow$)}
& \textcolor{green}{(57.06$\uparrow$)}
& \textcolor{red}{(2.31$\downarrow$)}
& \textcolor{green}{(5.69$\uparrow$)}
& \textcolor{red}{(1.21$\downarrow$)}
& \textcolor{green}{(66.49$\uparrow$)} \\
\rowcolor{darkgray!12}
&& ShadowFPT-II & 65.37 & \textbf{100.0} & 65.53 & \textbf{98.41} & 65.22 & \textbf{99.97} & 65.18 & \textbf{99.89} & 65.01 & \textbf{99.57} & 65.34 & \textbf{99.94} & 65.34 & \textbf{99.89} & 64.68 & \textbf{99.83} & 65.18 & \textbf{99.67} & 65.21 & \textbf{99.69} \\

&& & \textcolor{red}{(4.32$\downarrow$)}
& \textcolor{green}{(59.76$\uparrow$)}
& \textcolor{red}{(0.02$\downarrow$)}
& \textcolor{green}{(60.01$\uparrow$)}
& \textcolor{red}{(3.09$\downarrow$)}
& \textcolor{green}{(79.49$\uparrow$)}
& \textcolor{red}{(0.29$\downarrow$)}
& \textcolor{green}{(99.28$\uparrow$)}
& \textcolor{red}{(1.40$\downarrow$)}
& \textcolor{green}{(63.47$\uparrow$)}
& \textcolor{red}{(3.33$\downarrow$)}
& \textcolor{green}{(38.89$\uparrow$)}
& \textcolor{red}{(0.13$\downarrow$)}
& \textcolor{green}{(98.10$\uparrow$)}
& \textcolor{red}{(0.40$\downarrow$)}
& \textcolor{green}{(59.88$\uparrow$)}
& \textcolor{red}{(2.59$\downarrow$)}
& \textcolor{green}{(6.28$\uparrow$)}
& \textcolor{red}{(1.28$\downarrow$)}
& \textcolor{green}{(68.14$\uparrow$)} \\

\cmidrule(lr){2-23}

& \multirow{12}{*}{\rotatebox[origin=c]{90}{FPT-VL}}
& BadNets & 71.62 & 21.44 & 66.95 & 0.18 & 71.53 & 0.34 & \textbf{72.14} & 0.27 & 71.18 & 20.08 & 70.76 & 0.13 & 70.52 & 0.11 & 66.74 & 41.56 & 70.73 & 0.32 & \textbf{70.24} & 9.38 \\
&& Blended & 71.74 & 51.04 & 67.08 & 0.37 & 71.34 & 11.20 & 71.62 & 51.27 & 69.54 & 3.34 & \textbf{70.79} & 0.64 & 70.52 & 0.37 & 66.72 & 1.79 & 71.87 & 82.42 & 70.14 & 22.49 \\
&& FCBA & 71.61 & 0.91 & 66.98 & 0.13 & 71.66 & 0.34 & 71.13 & 0.32 & 69.99 & 0.37 & \textbf{70.79} & 0.16 & 70.61 & 0.13 & 66.93 & 0.45 & 70.44 & 0.72 & 70.02 & 0.39 \\
&& A3FL & 71.08 & 33.88 & 63.07 & 0.00 & 71.47 & 4.68 & 71.29 & 5.24 & 70.97 & 0.00 & 70.02 & 0.13 & 68.59 & 0.08 & 66.77 & 31.48 & 70.68 & 4.22 & 69.33 & 8.86 \\
&& Bad-PFL & 71.02 & 2.03 & \textbf{67.59} & 0.26 & 71.45 & 1.12 & 70.89 & 1.06 & 70.68 & 1.57 & 70.36 & 0.77 & 70.23 & 0.61 & 66.67 & 0.81 & 71.53 & 2.01 & 70.05 & 1.14 \\
&& PFedBA & \textbf{72.79} & 29.84 & 63.67 & 13.37 & \textbf{72.32} & 28.83 & 71.98 & 18.99 & \textbf{71.77} & 27.88 & 70.32 & 24.14 & 70.28 & 4.89 & 66.50 & 7.32 & \textbf{72.13} & 24.64 & 70.20 & 19.99 \\
&& Neurotoxin & 70.76 & 0.37 & 66.58 & 0.24 & 71.00 & 0.29 & 70.63 & 0.37 & 69.46 & 4.33 & 70.57 & 0.16 & 69.83 & 0.18 & \textbf{67.08} & 0.37 & 70.63 & 0.29 & 69.62 & 0.73 \\
&& CerP & 65.63 & 9.52 & 64.92 & 0.13 & 71.31 & 0.41 & 68.96 & 1.97 & 66.69 & 8.05 & 70.65 & 0.22 & \textbf{70.92} & 0.16 & 65.66 & 0.48 & 68.83 & 2.64 & 68.17 & 2.62 \\
\rowcolor{darkgray!12}
&& ShadowFPT-I & 69.54 & 99.70 & 63.78 & 97.03 & 69.44 & \textbf{99.62} & 69.07 & 99.62 & 70.05 & 99.78 & 67.22 & 98.98 & 68.42 & 99.35 & 64.36 & 99.67 & 69.83 & 99.78 & 67.97 & 99.28 \\
&& & \textcolor{red}{(3.25$\downarrow$)}
& \textcolor{green}{(48.66$\uparrow$)}
& \textcolor{red}{(3.81$\downarrow$)}
& \textcolor{green}{(83.66$\uparrow$)}
& \textcolor{red}{(2.88$\downarrow$)}
& \textcolor{green}{(70.79$\uparrow$)}
& \textcolor{red}{(3.07$\downarrow$)}
& \textcolor{green}{(48.35$\uparrow$)}
& \textcolor{red}{(1.72$\downarrow$)}
& \textcolor{green}{(71.90$\uparrow$)}
& \textcolor{red}{(3.57$\downarrow$)}
& \textcolor{green}{(74.84$\uparrow$)}
& \textcolor{red}{(2.50$\downarrow$)}
& \textcolor{green}{(94.46$\uparrow$)}
& \textcolor{red}{(2.72$\downarrow$)}
& \textcolor{green}{(58.11$\uparrow$)}
& \textcolor{red}{(2.30$\downarrow$)}
& \textcolor{green}{(17.36$\uparrow$)}
& \textcolor{red}{(2.27$\downarrow$)}
& \textcolor{green}{(76.79$\uparrow$)} \\

\rowcolor{darkgray!12}
&& ShadowFPT-II & 69.68 & \textbf{100.0} & 63.57 & \textbf{98.56} & 69.42 & 98.99 & 68.78 & \textbf{99.71} & 69.81 & \textbf{99.93} & 67.31 & \textbf{99.52} & 68.35 & \textbf{100.0} & 64.34 & \textbf{99.96} & 69.73 & \textbf{100} & 67.89 & \textbf{99.63} \\

&& & \textcolor{red}{(3.11$\downarrow$)}
& \textcolor{green}{(48.96$\uparrow$)}
& \textcolor{red}{(4.02$\downarrow$)}
& \textcolor{green}{(85.19$\uparrow$)}
& \textcolor{red}{(2.90$\downarrow$)}
& \textcolor{green}{(70.16$\uparrow$)}
& \textcolor{red}{(3.36$\downarrow$)}
& \textcolor{green}{(48.44$\uparrow$)}
& \textcolor{red}{(1.96$\downarrow$)}
& \textcolor{green}{(72.05$\uparrow$)}
& \textcolor{red}{(3.48$\downarrow$)}
& \textcolor{green}{(75.38$\uparrow$)}
& \textcolor{red}{(2.57$\downarrow$)}
& \textcolor{green}{(95.11$\uparrow$)}
& \textcolor{red}{(2.74$\downarrow$)}
& \textcolor{green}{(58.40$\uparrow$)}
& \textcolor{red}{(2.40$\downarrow$)}
& \textcolor{green}{(17.58$\uparrow$)}
& \textcolor{red}{(2.35$\downarrow$)}
& \textcolor{green}{(77.14$\uparrow$)} \\

\bottomrule
\end{tabular}}
\end{table}



\newpage
\section*{NeurIPS Paper Checklist}

\begin{enumerate}

\item {\bf Claims}
    \item[] Question: Do the main claims made in the abstract and introduction accurately reflect the paper's contributions and scope?
    \item[] Answer: \answerYes{}.
    \item[] Justification: The abstract and introduction clearly state the paper's scope, threat model, main methodological contribution, and empirical claims. The claims are supported by the experiments in Section~\ref{sec:experiments}, including evaluations across datasets, aggregation and defense rules, non-IID federated settings, and textual, visual, and joint vision--language prompt tuning.

\item {\bf Limitations}
    \item[] Question: Does the paper discuss the limitations of the work performed by the authors?
    \item[] Answer: \answerNo{}.
    \item[] Justification: The current version does not include a dedicated limitations discussion. Important limitations include the white-box threat model, the focus on CLIP-style frozen backbones, the use of one malicious client by default, and evaluation on three public image-recognition datasets.

\item {\bf Theory assumptions and proofs}
    \item[] Question: For each theoretical result, does the paper provide the full set of assumptions and a complete proof?
    \item[] Answer: \answerNA{}.
    \item[] Justification: The paper does not present formal theoretical results, theorems, or proofs. The equations in Section~\ref{sec:method} define the problem setup and optimization objectives rather than stating theoretical guarantees.

\item {\bf Experimental result reproducibility}
    \item[] Question: Does the paper fully disclose all the information needed to reproduce the main experimental results of the paper to the extent that it affects the main claims and/or conclusions of the paper?
    \item[] Answer: \answerYes{}.
    \item[] Justification: The paper specifies the main information needed to reproduce the experiments, including the CLIP backbone, datasets, federated setting, non-IID construction, prompt-tuning modalities, baselines, defenses, training details, trigger configuration, and evaluation metrics. Additional results and implementation details are provided in the appendix.

\item {\bf Open access to data and code}
    \item[] Question: Does the paper provide open access to the data and code, with sufficient instructions to faithfully reproduce the main experimental results, as described in supplemental material?
    \item[] Answer: \answerNo{}.
    \item[] Justification: The experiments use publicly available datasets and standard pretrained backbones, but the current submission does not provide an open-source code package or reproduction scripts. We plan to release anonymized code and detailed reproduction instructions when permitted by the submission policy.

\item {\bf Experimental setting/details}
    \item[] Question: Does the paper specify all the training and test details necessary to understand the results?
    \item[] Answer: \answerYes{}.
    \item[] Justification: The experimental setup specifies the backbone, datasets, non-IID client construction, number of clients, malicious-client setting, communication rounds, local epochs, batch size, learning rates, trigger configuration, attack baselines, defense baselines, and metrics. These details are provided in Section~\ref{sec:experiments} and the appendix.

\item {\bf Experiment statistical significance}
    \item[] Question: Does the paper report error bars suitably and correctly defined or other appropriate information about the statistical significance of the experiments?
    \item[] Answer: \answerNo{}.
    \item[] Justification: The paper reports results averaged over datasets, prompt-tuning settings, and aggregation or defense rules, but it does not report error bars, confidence intervals, or statistical significance tests over repeated random seeds. This is a limitation of the current empirical reporting.

\item {\bf Experiments compute resources}
    \item[] Question: For each experiment, does the paper provide sufficient information on the computer resources needed to reproduce the experiments?
    \item[] Answer: \answerNo{}.
    \item[] Justification: The current paper does not provide detailed compute-resource information such as GPU type, memory, runtime per experiment, or total compute budget. These details should be added in the appendix for full reproducibility.

\item {\bf Code of ethics}
    \item[] Question: Does the research conducted in the paper conform, in every respect, with the NeurIPS Code of Ethics?
    \item[] Answer: \answerYes{}.
    \item[] Justification: The research is conducted as a controlled security study using public datasets and standard experimental settings. The paper analyzes a dual-use backdoor threat to improve understanding of vulnerabilities in federated prompt tuning and does not involve unauthorized system access, private user data, or human-subject data collection.

\item {\bf Broader impacts}
    \item[] Question: Does the paper discuss both potential positive societal impacts and negative societal impacts of the work performed?
    \item[] Answer: \answerYes{}.
    \item[] Justification: The work has both positive and negative societal implications. It exposes a realistic security risk in federated prompt tuning, which can motivate stronger defenses, but the proposed attack is dual-use and could be misused if applied outside controlled research settings.

\item {\bf Safeguards}
    \item[] Question: Does the paper describe safeguards that have been put in place for responsible release of data or models that have a high risk for misuse?
    \item[] Answer: \answerNA{}.
    \item[] Justification: The paper does not release a high-risk pretrained model, image generator, scraped dataset, or deployable system. The study uses public datasets and evaluates attacks in an offline experimental setting.

\item {\bf Licenses for existing assets}
    \item[] Question: Are the creators or original owners of assets used in the paper properly credited, and are the license and terms of use explicitly mentioned and properly respected?
    \item[] Answer: \answerYes{}.
    \item[] Justification: The paper cites the original sources for the existing datasets, pretrained CLIP backbone, prompt-tuning methods, attack baselines, and defense baselines used in the experiments. The experiments are based on publicly available research assets.

\item {\bf New assets}
    \item[] Question: Are new assets introduced in the paper well documented, and is the documentation provided alongside the assets?
    \item[] Answer: \answerNA{}.
    \item[] Justification: The paper does not introduce a new dataset, benchmark, pretrained model, or other standalone research asset. The contribution is a method and empirical security analysis.

\item {\bf Crowdsourcing and research with human subjects}
    \item[] Question: For crowdsourcing experiments and research with human subjects, does the paper include the full text of instructions given to participants and screenshots, if applicable, as well as details about compensation?
    \item[] Answer: \answerNA{}.
    \item[] Justification: The paper does not involve crowdsourcing, user studies, human-subject experiments, or participant compensation. All experiments are conducted on existing public image-recognition datasets.

\item {\bf Institutional review board approvals or equivalent for research with human subjects}
    \item[] Question: Does the paper describe potential risks incurred by study participants, whether such risks were disclosed to the subjects, and whether Institutional Review Board approvals or an equivalent approval/review were obtained?
    \item[] Answer: \answerNA{}.
    \item[] Justification: The paper does not involve human subjects, crowdsourcing, private user data, or participant interaction. Therefore, IRB approval or equivalent human-subjects review is not applicable.

\item {\bf Declaration of LLM usage}
    \item[] Question: Does the paper describe the usage of LLMs if it is an important, original, or non-standard component of the core methods in this research?
    \item[] Answer: \answerNA{}.
    \item[] Justification: The core method, experiments, and scientific claims do not rely on LLMs as an important, original, or non-standard component. Any use of LLMs for writing, editing, or formatting does not affect the methodology, experimental design, or scientific conclusions.

\end{enumerate}

\end{document}